\documentclass[conference]{IEEEtran}
\usepackage{times}

\usepackage[numbers]{natbib}
\usepackage{multicol}
\usepackage[bookmarks=true,hidelinks]{hyperref}
\usepackage{url}
\usepackage{graphicx} 
\usepackage{subcaption}
\usepackage{wrapfig}
\usepackage{booktabs}
\usepackage{multirow}
\usepackage{amsfonts}
\usepackage{listings}
\usepackage{xcolor} 
\usepackage{comment}
\usepackage{amsmath}
\usepackage{amssymb}

\lstdefinestyle{mystyle}{
    backgroundcolor=\color{black!5},
    commentstyle=\color{green!40!black},
    keywordstyle=\color{blue},
    numberstyle=\tiny\color{black!50},
    stringstyle=\color{purple},
    basicstyle=\ttfamily\footnotesize,
    breakatwhitespace=false,
    breaklines=true,
    captionpos=b,
    keepspaces=true,
    numbers=left,
    numbersep=5pt,
    showspaces=false,
    showstringspaces=false,
    showtabs=false,
    tabsize=2
}

\begin{document}

\title{CoRAL: Contact-Rich Adaptive LLM-based Control for Robotic Manipulation}

\author{
\authorblockN{
Berk Cicek$^{*}$,
Mert K. Er$^{*}$,
Ozgur S. Oguz
}
\authorblockA{
LiRA Lab, Department of Computer Engineering\\
Bilkent University, Ankara, T\"urkiye\\
Emails: \{berk.cicek, kaan.er, ozgur.oguz\}@bilkent.edu.tr\\
$^{*}$These authors contributed equally to this work.
}
}

\maketitle

\begin{abstract}
While Large Language Models (LLMs) and Vision-Language Models (VLMs) demonstrate remarkable capabilities in high-level reasoning and semantic understanding, applying them directly to contact-rich manipulation remains a challenge due to their lack of explicit physical grounding and inability to perform adaptive control.
To bridge this gap, we propose CoRAL (Contact-Rich Adaptive LLM-based control), a modular framework that enables zero-shot planning by decoupling high-level reasoning from low-level control. 
Unlike black-box policies, CoRAL utilizes Large Language Models (LLMs) not as direct controllers, but as cost designers that synthesize context-aware objective functions for a sampling-based motion planner (MPPI). 
To address the ambiguity of physical parameters in visual data, we introduce a neuro-symbolic adaptation loop: a Vision-Language Model provides semantic priors for environmental dynamics (e.g., mass, friction estimates), which are then explicitly refined in real-time via online system identification, while the LLM iteratively modulates the cost function structure to correct strategic errors based on interaction feedback. 
Furthermore, a retrieval-based memory unit allows the system to reuse successful strategies across recurrent tasks. 
This hierarchical architecture ensures real-time control stability by decoupling high-level semantic reasoning from reactive execution, effectively bridging the gap between slow LLM inference and dynamic contact requirements. 
We validate CoRAL on both simulation and real-world hardware across challenging and novel tasks, such as ``flipping objects against walls" leveraging extrinsic contacts. 
Experiments demonstrate that CoRAL outperforms state-of-the-art VLA and foundation-model-based planner baselines by boosting success rates over 50\% on average in unseen contact-rich scenarios, effectively handling sim-to-real gaps through its adaptive physical understanding. 
Website: \textit{\href{https://sites.google.com/view/lira-coral}{https://sites.google.com/view/lira-coral}}
\end{abstract}

\IEEEpeerreviewmaketitle

\section{Introduction}
\begin{figure}[t]
\begin{center}
\includegraphics[width=1\linewidth]{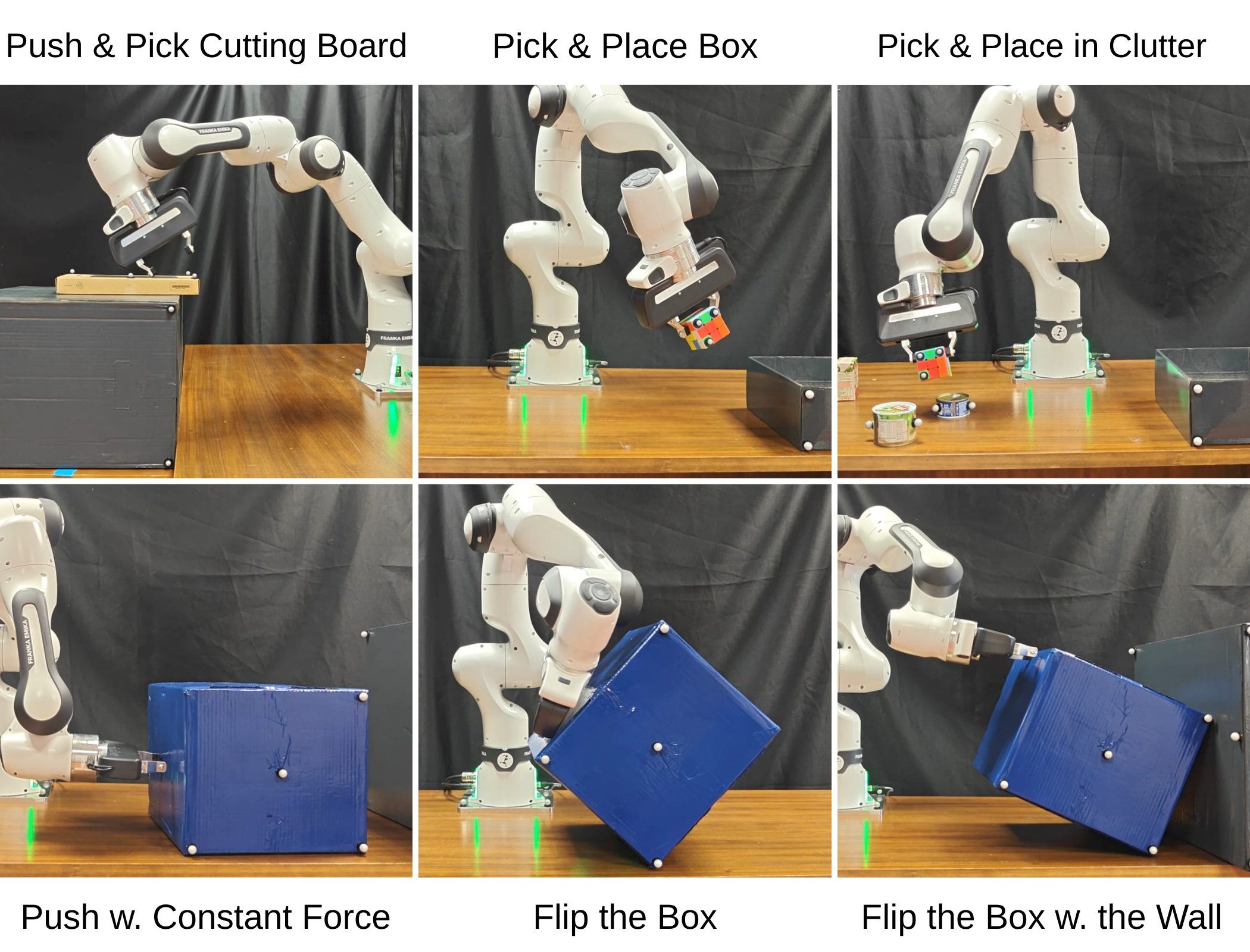}
\end{center}
\caption{Real-world execution of CoRAL across six different manipulation tasks.}
\label{fig:real_robot}
\vspace{-15pt}
\end{figure}

Foundational models have demonstrated significant success in various fields, leading to increased efforts to apply these models within robotics~\citep{firoozi2025foundation,tayyab2025foundation}. 
Typically, these models are integrated into robotic control pipelines through \textbf{imitation learning}, most notably in the form of Vision-Language-Action (VLA) systems~\citep{ma2024survey,zhong2025survey,sapkota2025vision}.
However, existing VLA frameworks struggle to effectively handle contact-rich manipulation tasks, which constitute a substantial portion of daily interactions~\citep{hao2025tla,yu2025forcevla,xue2025reactive}. 
In this work, we specifically refer to a challenging subset of these tasks as contact- and force-critical, where success depends not merely on geometric collision avoidance, but on the precise, active regulation of extrinsic contact forces to manipulate objects (e.g., pivoting against a wall - Fig.~\ref{fig:real_robot}).
These tasks pose significant challenges, as they require not only precise trajectory planning but also sophisticated interaction force management and adaptive control strategies. 
Achieving success in such complex scenarios typically necessitates extensive training through teleoperation or detailed dynamic modeling, methods that are labor-intensive and reduce generalizability.

\begin{figure*}[t]
\centering
\includegraphics[width=0.8\textwidth]{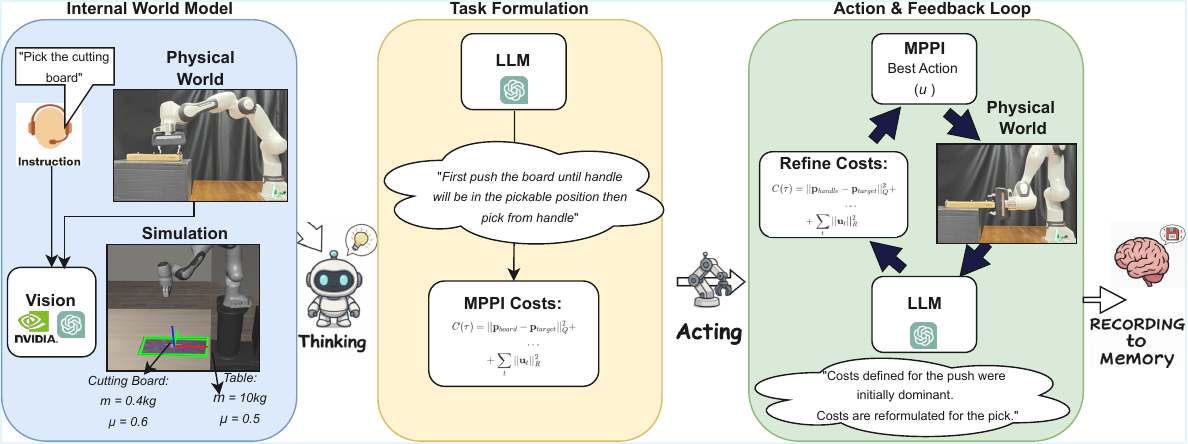}
\caption{
The conceptual workflow of CoRAL, illustrated with the ``pick the cutting board" task.}
\label{fig:conceptual_flow}
\vspace{-15pt}
\end{figure*}

Humans, by contrast, rely on initial estimations, subsequently refine their strategies based on sensory feedback, and adjust interactions accordingly~\citep{flanagan2006control,johansson2009coding,kim2015multimodal}. 
Similar to this cognitive framework, we propose a novel modular system, \textbf{Contact-Rich Adaptive LLM-based Control (CoRAL)}, that integrates reasoning, planning, and control modules into a cohesive architecture. 
Our model begins by estimating 6-DoF object poses from RGB-D data using FoundationPose~\cite{wen2024foundationpose}, and then a VLM generates \textit{semantic beliefs} regarding physical parameters such as mass and friction from the estimated object poses, the environment image, and the textual task description (Fig.~\ref{fig:conceptual_flow}).
The planning stage generates initial contact strategies and actions, which are executed in the evaluation environment through reactive control modules. 
The outcomes from these actions are continuously monitored, with the interaction history and control feedback being used for iterative refinement of plans.

A key innovation of our approach is the strategic integration of vision and Large Language Models (LLMs) with motion planners and controllers, substantially enhancing action explainability and enabling more effective reasoning. 
Our modular structure clearly delineates roles: vision module manages parameter estimation for environment modeling, while LLM provides symbolic reasoning, initial contact strategies, and cost estimations. 
The reactive controller then applies these symbolic outputs in the evaluation world, establishing a tight feedback loop between high-level strategy and low-level sensory information.
Our main contributions are:
\begin{itemize}
\item We propose a novel framework that employs Large Language Models (LLMs) not as direct controllers, but as context-aware \textit{cost designers} for a high-frequency motion planner, enabling zero-shot planning for dynamic, contact- and force-critical manipulation.

We introduce a modular architecture that decouples perception (providing semantic physics priors) from reasoning (formulating control objectives). This separation offers a transparent and adaptive alternative to monolithic VLA architectures, allowing the system to maintain robustness even when initial visual estimates are imperfect.

\item We develop a hierarchical closed-loop mechanism where reactive control ensures contact stability, while a slower LLM-driven loop performs online system identification to adapt physical parameter beliefs and strategy mid-execution.

\item Our framework's long-term adaptability is enhanced by a retrieval-based memory unit that stores successful physical parameters and contact strategies to bootstrap effective solutions for novel, yet semantically similar tasks.
\end{itemize}

We evaluate CoRAL on a challenging suite of manipulation tasks, including novel contact- and force-critical problems, such as picking up a thin object from a table and standardized benchmarks from the \textsc{LIBERO} suite~\cite{liu2023libero}. 
Furthermore, we validate the system's robustness against real-world physical uncertainties by deploying it on a physical robot.
Our experiments and detailed ablation studies confirm that this modular structure offers a robust, data-efficient alternative for contact-rich manipulation, providing explainability and adaptability in zero-shot regimes where end-to-end models struggle.

\section{Related Work}
\label{sec:related_work}
\textbf{From End-to-End Policies to Decoupled Reasoning.}
Foundation models have shifted robotic manipulation towards VLA models learning general-purpose policies~\cite{firoozi2025foundation, ma2024survey}. Leading examples like OpenVLA~\cite{kim2025openvla}, $\pi_0$~\cite{black2024pi_0}, and RT-X~\cite{o2024open} directly map multimodal inputs to low-level actions. While powerful, their reliance on imitation learning makes them data-dependent and often brittle in novel physical scenarios involving complex contact dynamics. To overcome this, emerging frameworks decouple high-level reasoning from low-level control. ThinkAct~\cite{huang2025thinkact}, Inner Monologue~\cite{huang2022innermonologue}, and ECoT~\cite{zawalski2024ecot} leverage LLMs to generate reasoning steps guiding separate, learned policies. Similarly, MolmoAct~\cite{lee2025molmoact} produces mid-level spatial plans, while OneTwoVLA~\cite{lin2025onetwovla} formalizes this as System 1 (acting) and System 2 (reasoning). CoRAL aligns with this decoupling but takes a distinct neuro-symbolic path: we ground LLM reasoning directly in a controller rather than a learned model.

\textbf{Integrating Foundation Models with Motion Planners and Controllers.}
 Alternatively, foundation models can guide traditional motion planners using semantic understanding. VoxPoser~\cite{huang2023voxposer} and IMPACT~\cite{ling2025impact} use VLMs to generate static 3D cost maps for planners like MPPI or RRT*, while VLMPC~\cite{zhao2024vlmpc} embeds a VLM within an MPC~\cite{garcia1989model} loop. In code generation, Eureka~\cite{ma2023eureka} and DrEureka~\cite{ma2024dreureka} synthesize RL reward functions but operate \textit{offline}. Closest to our approach, Language-to-Rewards (L2R)~\cite{yu2023language} generates rewards for \textit{real-time} MPC. However, L2R relies on static physical assumptions and lacks mechanisms to correct model mismatches during execution. CoRAL advances this by elevating the LLM to a high-level strategist capable of online adaptation. Rather than merely identifying goals, our LLM formulates the structure of the MPPI~\cite{williams2017model} cost function and symbolic contact strategies, grounding commonsense reasoning directly into the optimal control problem.

\textbf{Tackling Contact-Rich Manipulation.}
Contact-rich manipulation requires nuanced force control beyond simple trajectory generation. Recent works like ForceVLA~\cite{yu2025forcevla}, TLA~\cite{hao2025tla}, VLA-Touch~\cite{bi2025vlatouch}, RDP~\cite{xue2025reactive}, and FACTR~\cite{liu2025factr} explicitly integrate force or tactile data into learned policies. While effective, this hardware-centric approach creates a data bottleneck, requiring difficult-to-collect specialized multimodal datasets~\cite{firoozi2025foundation}. CoRAL leverages real-time force feedback within the MPPI controller but eliminates the need for prior demonstration datasets. We use the LLM to formulate high-level strategies and cost functions explicitly reasoning about forces, which the controller executes adaptively. This neuro-symbolic approach combines physical sensing with zero-shot reasoning, avoiding the imitation learning bottleneck while achieving precise, force-aware control.

\begin{figure*}[t]
\centering
\includegraphics[width=\textwidth]{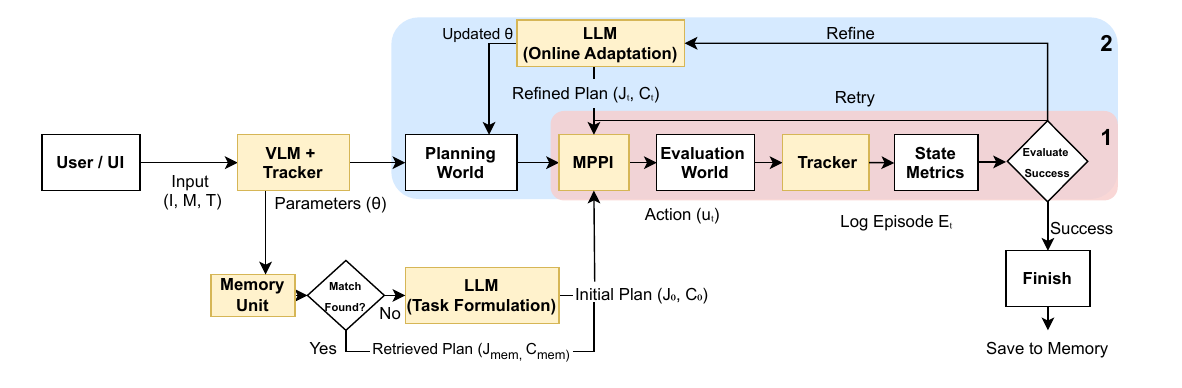}
\caption{
The overall architecture of the \textbf{CoRAL} framework. 
Given an input image $I$, object models $M$ and task description $T$, the vision module extracts world parameters $\theta$. 
If the Memory Unit finds a similar successful experience, its retrieved plan ($J_{\textrm{mem}}, C_{\textrm{mem}}$) is used to guide the MPPI controller. Otherwise, the LLM (Task Formulation) module generates an initial plan ($J_0, C_0$). 
The system then enters the main execution cycle, which is governed by two nested feedback loops labeled (1) and (2).
\textbf{(1) The Inner Loop} is a high-frequency re-planning cycle. At each step, the MPPI, guided by the current plan, generates an action $u_t$ based on the latest `State Metrics'. This loop (`Retry') continues until the task succeeds or a refinement is needed.
\textbf{(2) The Outer Loop} is a low-frequency, high-level adaptation cycle. If the inner loop fails persistently, the `Refine' path is taken, where the LLM (Online Adaptation) updates both the world model parameters ($\theta$) for the `Planning World' and the strategic `Refined Plan ($J_t, C_t$)' for the MPPI. 
Successful episodes are stored back into the Memory Unit.
}
\vspace{-10pt}
\label{fig:method_flow}
\end{figure*}

\section{Methodology}
\label{sec:method}

CoRAL is a neuro-symbolic framework designed for zero-shot, contact-rich manipulation. It strategically decouples high-level reasoning from low-level control by integrating a vision pipeline that continuously tracks object poses and enriches the world model with semantic physics priors inferred by the VLM, an LLM acting in two distinct roles (Task Formulation and Online Adaptation), a Memory Unit for experience retrieval, and a Model Predictive Path Integral controller (MPPI) for reactive execution. The overall architecture, which features nested feedback loops for rapid and robust adaptation, is illustrated in Figure~\ref{fig:method_flow}. Below, we detail each component of this architecture.

\subsection{Environment Perception and World Model Initialization}

The first step is to translate raw visual, textual, and geometric inputs into a structured, physics-aware world model. Our perception pipeline achieves this through a two-stage process that first establishes the geometric state of the scene and then initializes it with semantic physical beliefs. The process involves two core steps:
\begin{enumerate}
    \item \textbf{Pose Estimation and Tracking:} We employ \textbf{FoundationPose}~\cite{wen2024foundationpose}, a state-of-the-art pose estimation model, to determine and continuously track the 6-DoF poses of all interactable objects. This model takes the RGB-D camera images $I$, and the known 3D geometric models of the objects, $M$, as input. The output is a real-time stream of estimated pose data for each object in the scene.

    \item \textbf{Semantic Physics Priors:} Direct estimation of physical dynamics from static images is ill-posed. Therefore, we utilize the VLM not as a ground-truth estimator, but to generate \textit{semantic priors}. Given the visual input and task context $T$, the VLM infers likely physical properties (e.g., distinguishing a "heavy metal tool" from a "light foam block"). These priors initialize the mass and friction coefficients in $\theta$, which are explicitly treated as beliefs subject to refinement by the online adaptation module.
\end{enumerate}

The combined output of this perception pipeline is a structured set of world parameters, $\theta$. For each object, $\theta$ contains its semantic label (derived from the input 3D model), its continuously tracked pose from FoundationPose, and its semantic physics priors from the VLM. These parameters are crucial as they are used to initialize and continuously update the internal \texttt{Planning World} that the MPPI planner operates on.

While our current implementation focuses on mass and friction—the 
dominant parameters for rigid-body contact dynamics in our evaluated 
tasks—the framework is intentionally designed to be parameter-agnostic. 
The VLM prompt and JSON schema (detailed in supplementary material) 
can be extended to query additional properties such as stiffness for 
soft objects, damping coefficients for viscous interactions, or 
center-of-mass offsets for asymmetric objects. Such properties can 
be incorporated into MPPI's rollout dynamics without architectural 
changes, enabling broader deployment beyond rigid-body manipulation.

\subsection{LLM-driven Task Formulation and Memory Retrieval}

With the perceived world, the system formulates a concrete plan. This is handled by the `LLM (Task Formulation)' module, which can generate a plan from scratch or leverage past experiences from the `Memory Unit'.

\textbf{Memory Retrieval:} Before invoking the LLM, the system queries the `Memory Unit' with the current world parameters $\theta$ and natural language task description $T$. Our memory module is based on Retrieval-Augmented Generation (RAG), storing successful ``experience episodes" indexed by task definitions and environmental parameters. Instead of relying on predefined similarity metrics, the LLM embeds the current task into a latent semantic space to retrieve the most relevant past experience:
\begin{equation}
    (J_{\textrm{mem}}, C_{\textrm{mem}}) = \text{RAG}_{\text{Retrieve}}(T, \theta)
\end{equation}
where $J_{\textrm{mem}}$ denotes the final cost function that led to a successful episode, and $C_{\textrm{mem}}$ denotes the corresponding contact strategy.
If a sufficiently similar and successful past experience is found, its stored plan ($J_{\textrm{mem}}, C_{\textrm{mem}}$) is retrieved and used as the initial plan, bypassing the initial LLM call and accelerating performance.

\textbf{Plan Generation from Scratch:} If no suitable memory is found, the `LLM (Task Formulation)' module is invoked. It acts as a high-level strategist, translating the task $T$ and world parameters $\theta$ into a formal optimization problem. Its output is an initial plan tuple $(J_0, C_0)$, where:
\begin{itemize}
    \item \textbf{Initial MPPI Cost Function ($J_0$):} The LLM generates the mathematical structure and relative weights of a cost function. Specifically, for a given task, the LLM provides a structured cost functional, for instance:
    \begin{align}
    \label{eq:cost_function}
    J_0(\mathbf{x}_{0:H}, \mathbf{u}_{0:H-1}) = \sum_{t=0}^{H-1} \Bigl[ & w_d\,\bigl\lVert \mathbf{p}_{\text{target}} - \mathbf{p}_{\text{obj}}(t)\bigr\rVert^2 \nonumber \\
    & +\; w_c\,\mathbb{I}\{\text{no contact at }t\} \nonumber \\
    & +\; w_u\,\lVert \mathbf{u}_t\rVert^2 \Bigr]
    \end{align}
    Here, the weights $w_d, w_c, w_u$ and the cost terms are determined by the LLM based on the task description (e.g., for a pushing task, $w_c$ would be high). In the example cost function, $\mathbf{p}_{\text{obj}}(t)$ is the object's tracked position at time $t$, and $\mathbb{I}\{\cdot\}$ is an indicator function penalizing the absence of contact. This expression is only an illustrative example: in general, the LLM is free to introduce any cost terms constructible from the available state, pose, and action variables, and is not restricted to a fixed finite set of cost terms. Moreover, the LLM can solve the task in stages by defining separate cost functions and specifying conditions for transitioning between stages.

\item \textbf{Initial Contact Strategy ($C_0$):} To bridge the gap between abstract task descriptions and continuous geometry, we employ the LLM as a \textit{high-level semantic heuristic} to predict a \textit{coarse region of interest} from the task description. The LLM outputs a nominal reference point $c_{\mathrm{ref}}$ in the object's local frame and a spatial expansion factor $r$. A procedural helper then constructs an ellipsoid $\mathcal{E}(c_{\mathrm{ref}}, \mathbf{A})$ centered at $c_{\mathrm{ref}}$, with axes $\mathbf{A}$ aligned with the object's principal geometric axes and scaled by $r$. Crucially, this expansion serves as a spatial buffer that compensates for the imprecision of LLM-generated coordinates: even if $c_{\mathrm{ref}}$ deviates from the ideal contact location, $r$ ensures the valid contact manifold $\mathcal{R}$ still covers the optimal region. The valid contact manifold is defined as $\mathcal{R} \;=\; \partial\mathcal{O} \cap \mathcal{E}(c_{\textrm{ref}}, \mathbf{A}).$ We then sample candidate target points from this manifold to form the strategy set: $C_0 \;=\; \Bigl\{ x_{\textrm{des}} \mid x_{\textrm{des}} \sim \mathrm{Unif}\bigl(\mathcal{R}\bigr) \Bigr\}.$ Finally, a target $x_{\textrm{des}} \in C_0$ is integrated into the optimization not as a hard constraint, but as a \textit{soft cost attractor} in $J_0$ (e.g., $w_{\textrm{attr}}\|p_{\textrm{eef}} - x_{\textrm{des}}\|^2$). This formulation creates a “stretched” cost landscape: it biases MPPI sampling toward the semantic region of interest while, due to the soft penalty, still allowing deviation to find the dynamically optimal contact point within that vicinity, improving robustness to perception noise.
\end{itemize}

\subsection{Reactive Planning and Execution (The Inner Loop)}
The core of our system is a high-frequency, reactive execution cycle governed by the MPPI controller. This corresponds to the Inner Loop (1) in Figure~\ref{fig:method_flow}.
To address the latency challenges inherent in LLM inference, we implement a strict hierarchical control architecture:
\begin{itemize}
    \item \textbf{Tier 1 (Hardware Level):} Joint impedance control running at 1kHz guarantees safety and compliance during physical interaction.
    \item \textbf{Tier 2 (Trajectory Level):} The MPPI planner runs at 10Hz, sampling trajectories based on the current cost function $J_t$.
    \item \textbf{Tier 3 (Reasoning Level):} The LLM operates asynchronously ($\sim$1Hz) to update the cost structure periodically.
\end{itemize}
\textbf{MPPI Formulation:} The MPPI controller solves a stochastic optimal control problem at each timestep. Given a state-transition model $x_{t+1} = f(x_t, u_t) + \epsilon_t$, where $x_t$ is the state, $u_t$ is the target end-effector delta pose, and $\epsilon_t$ is system noise, the objective is to find $U = \{u_0, \dots, u_{H-1}\}$ minimizing the expected total cost:
\begin{equation}
\label{eq:mppi_objective}
    U^* = \arg\min_{U} \mathbb{E}\left[\phi(x_{H}) + \sum_{t=0}^{H-1}q(x_t, u_t)\right]
\end{equation}
where $\phi(x_H)$ is a terminal state cost and $q(x_t, u_t)$ is the running cost at each step, directly defined by the LLM-generated cost function $J_0$ (see Eq.~\ref{eq:cost_function} for an example structure).
MPPI approximates this optimization by:
\begin{enumerate}
    \item Sampling $K$ control sequence perturbations $\delta U_k \sim \mathcal{N}(0, \Sigma)$ from a Gaussian distribution.
    \item Creating $K$ rollout trajectories by applying the perturbed control sequences $V_k = U_{prev} + \delta U_k$ in the `Planning World'.
    \item Calculating the total cost $S(V_k)$ for each of the $K$ trajectories.
    \item Computing an exponentially weighted average of the perturbations to update the control sequence:
    \begin{equation}
        \begin{split}
            U_{new} &= U_{prev} + \sum_{k=1}^{K} w_k \delta U_k, \\
            \text{where} \quad w_k &= \frac{\exp\left(-\frac{1}{\lambda} S(V_k)\right)}{\sum_{j=1}^{K} \exp\left(-\frac{1}{\lambda} S(V_j)\right)}
        \end{split}
    \end{equation}
\end{enumerate}
Following the receding horizon principle, only the first action, $u_0$, of the newly optimized sequence $U_{new}$ is executed.

\textbf{Reactive Control Augmentation:} To achieve robustness against the inherent sim-to-real gap, we augment the nominal planned action with a real-time feedback term. The final control command $\nu_t$ sent to the robot is:
\begin{equation}
    \nu_t = u_t + K_f \cdot (x_{\text{des}} - x_{\text{measured}})
\end{equation}
where $u_t$ is the action computed by MPPI, the error term is calculated from real-time sensors (e.g., force/torque, proprioception), and $K_f$ is a feedback gain matrix. This `Retry' loop continues at a high frequency, constantly re-planning and correcting based on physical feedback.

\subsection{Online Adaptation via LLM-driven Refinement (The Outer Loop)}

After a predefined number of attempts, a hyperparameter we denote as $N_{\textrm{retry}}$, the system triggers the low-frequency Outer Loop (2). This invokes the `LLM (Online Adaptation)' module, which acts as a diagnostician and re-strategist.

The input to this module is the logged episode data $E_t$, which contains the history of states, actions, the contact strategies and cost functions that were used, and the estimated physical parameters that led to the failure. By analyzing this rich context, the LLM performs two critical functions:
\begin{enumerate}
    \item \textbf{World Model Correction:} Acting as a system identification agent, the LLM refines the semantic priors. For example, if the robot pushes an object but the object moves less than predicted, the LLM can infer that its initial estimate of the object's mass was too low and output an `Updated $\theta$'.
    \item \textbf{Strategy Refinement:} The LLM can also alter the plan itself. It might change the weights of the cost function (e.g., prioritizing force control over position accuracy) or propose an entirely new contact strategy. This results in a `Refined Plan ($J_t, C_t$)'.
\end{enumerate}
This refined world model and plan are then fed back into the inner loop, allowing the system to learn from its failures and adapt its entire approach within a single task execution.

\section{Experiments}
\label{sec:experiments}

We conducted a series of experiments in a simulated environment, complemented by a real-world validation study, to rigorously evaluate the performance of CoRAL. Our evaluation is designed to answer four key research questions: 
\textbf{(RQ1)} How does CoRAL perform on complex, contact-rich manipulation tasks in a zero-shot setting compared to state-of-the-art baselines? 
\textbf{(RQ2)} How critical is each core component of our neuro-symbolic architecture—specifically the vision/language model role separation, the online refinement loop, and the memory unit—to the overall success? 
\textbf{(RQ3)} Can the system demonstrate robustness and adaptability by reasoning about and recovering from failures?
\textbf{(RQ4)} Can CoRAL's performance be effectively carried to a real system?

\textbf{Testing Environments:} The simulated experiments were conducted in the evaluation world, implemented using \textsc{robosuite} library~\cite{zhu2020robosuite}, which is based on the \textsc{MuJoCo} physics engine~\cite{todorov2012mujoco}. The robot is a simulated 7-DoF Franka Emika Panda arm with a parallel-jaw gripper. Sensory inputs include RGB-D images from a fixed camera, proprioceptive feedback, and force/torque data, which are provided by the robot's simulated sensors. In addition to our custom environments, two benchmark tasks from the \textsc{LIBERO} suite~\cite{liu2023libero} were also incorporated for evaluation.

We additionally evaluate on a real Franka Emika Panda using the same sensing modalities (fixed-view RGB-D, proprioception, and force/torque), where force/torque readings are obtained from the robot's built-in actuator sensors. For real-world state estimation, we replace \textsc{FoundationPose} with a motion capture setup for simplicity.

The implementation details are included in supplementary material.

\begin{figure}[h]
\begin{center}
\includegraphics[width=1\linewidth]{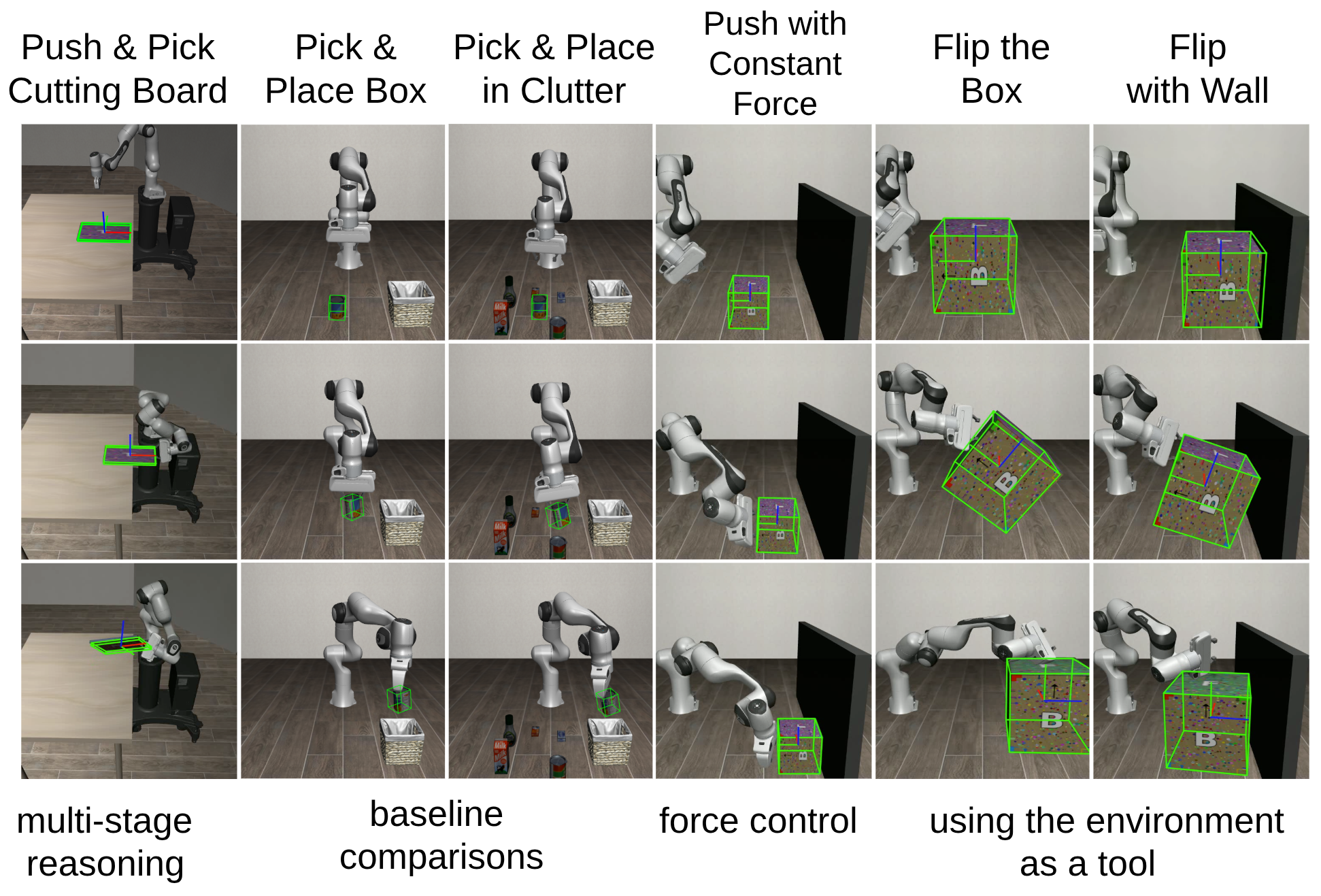}
\end{center}
\caption{CoRAL on six different tasks with the tracked pose overlay of the object of interest.}
\label{fig:tasks}
\end{figure}

\paragraph{Tasks and Evaluation Metrics}
We evaluated our framework on four challenging, contact- and force-critical manipulation tasks and two standard pick and place tasks, shown in Fig.~\ref{fig:tasks}, designed to be difficult for purely vision-based, collision-avoidant planners. Each task was performed 10 times with randomized initial object poses, object masses, surface friction coefficients and the object dimensions for the box and the board objects. The tasks are as follows:
\textbf{T1: Push and Pick Cutting Board}, a multi-stage task testing pushing and reasoning about object parts and pose for grasping;
\textbf{T2: Pick Box \& T3: Pick and Place in Clutter}, standard pick-and-place tasks to establish a baseline;
\textbf{T4: Push with Constant Force}, testing the reactive controller's ability to manage force feedback;
\textbf{T5: Flip Box}, a dynamically complex maneuver;
and \textbf{T6: Flip with Wall}, requiring multi-contact reasoning to use the wall as a fixture.
We primarily use Success Rate (binary measure across 10 trials) to evaluate performance.

\label{app:tasks}

\paragraph{Comparative Baselines}
We compare CoRAL against three state-of-the-art methods and four internal ablations.
The \textbf{State-of-the-Art Baselines} consist of two categories:
\textbf{1) End-to-End VLA Models:} We evaluate \textbf{OpenVLA-OFT}~\cite{kim2025fine} and \textbf{$\pi_{0.5}$}~\cite{intelligence2504pi0}. For these models, we rely on the officially released \textsc{LIBERO}-\textsc{Object} checkpoint for pick-and-place tasks and the \textsc{LIBERO}-\textsc{Goal} checkpoint for all other tasks. This setup tests CoRAL's zero-shot capabilities against powerful, pre-trained policies.
\textbf{2) Cost-Generation Baseline:} We evaluate \textbf{L2R}~\cite{yu2023language}, a method that uses LLMs to synthesize cost functions for an MPC controller. This baseline serves as a direct comparison for our neuro-symbolic planning approach but without the online adaptation mechanism.
 In addition, we include two \textbf{Human Expert-Designed Cost} baselines. In the \emph{single-stage} variant, an expert manually designs a single MPPI running-cost for each task. In the \emph{FSM} variant, the expert is allowed to construct an explicit finite-state machine with phase-specific costs (e.g., push--then--pick or push--then--flip). In both cases, the cost functions are tuned in a separate design environment and then evaluated \emph{as-is} in our randomized test environment, providing an upper bound on what carefully engineered, task-specific objectives can achieve.
Our \textbf{Ablation Baselines} are:
\textbf{CoRAL (w/o Pose Tracking)}, which removes FoundationPose and relies on the VLM to estimate object poses, testing the criticality of a dedicated pose estimator;
\textbf{CoRAL (w/o Memory)}, which removes the experience retrieval mechanism;
\textbf{CoRAL (w/o Refinement)}, which disables the online adaptation loop;
and \textbf{CoRAL (Unified VLM)}, which uses a single multimodal prompt for both perception and planning to test the importance of separating VLM/LLM roles.

\begin{table*}[!t]
\centering
\caption{Comparison against the baselines and ablation study. Performance is measured by success rate (x/10 trials).} 
\label{tab:comprehensive_results}
\resizebox{\textwidth}{!}{%
\begin{tabular}{l c c c c c c} 
\toprule
\textbf{Method} & \textbf{T1: Push+Pick} & \textbf{T2: Pick+Place} & \textbf{T3: Clutter} & \textbf{T4: Const. Force} & \textbf{T5: Flip Box} & \textbf{T6: Flip w/ Wall} \\
\midrule
\multicolumn{7}{l}{\textit{State-of-the-Art Baseline}} \\
\midrule
OpenVLA-OFT~\cite{kim2025fine} & 0/10 & 10/10 & 9/10 & 0/10 & 1/10 & 0/10 \\
$\pi_{0.5}$~\cite{intelligence2504pi0} & 0/10 & 10/10 & 8/10 & 0/10 & 3/10 & 0/10 \\
L2R ~\cite{yu2023language} & 0/10 & 10/10 & 9/10 & 5/10 & 4/10 & 1/10 \\
\midrule
\multicolumn{7}{l}{\textit{Human Expert-Designed Cost Baselines}} \\
\midrule
Expert (single-stage) & 0/10 & 10/10 & 10/10 & 9/10 & 9/10 & 3/10 \\
Expert (FSM) & 8/10 & 10/10 & 10/10 & 10/10 & 10/10 & 9/10 \\
\midrule
\multicolumn{7}{l}{\textit{Our Method (Ablation Study)}} \\
\midrule
\textbf{CoRAL (Ours)} & \textbf{5/10} & \textbf{10/10} & \textbf{10/10} & \textbf{9/10} & \textbf{9/10} & \textbf{7/10} \\
CoRAL (w/o Memory) & 2/10 & 10/10 & 9/10 & 9/10 & 7/10 & 5/10 \\
CoRAL (w/o Refinement) & 0/10 & 10/10 & 3/10 & 6/10 & 4/10 & 2/10 \\
CoRAL (Unified VLM) & 0/10 & 2/10 & 0/10 & 1/10 & 0/10 & 0/10 \\
CoRAL (w/o Pose Tracking) & 0/10 & 0/10 & 0/10 & 0/10 & 0/10 & 0/10 \\
\bottomrule
\end{tabular}%
}
\end{table*}

\subsection{Results and Analysis}
Table~\ref{tab:comprehensive_results} presents a comprehensive overview of our experimental findings.

\subsubsection{State-of-the-Art Comparison (RQ1)}
CoRAL significantly outperforms both state-of-the-art VLA baselines (OpenVLA-OFT, $\pi_{0.5}$) and the foundation-model-based planner baseline (L2R), particularly in tasks requiring sophisticated physical reasoning (T1, T4, T5, T6). While all baselines perform well on the simpler pick-and-place tasks (T2, T3), their performance degrades sharply on the more complex, contact-rich scenarios. This highlights distinct limitations in existing approaches. On one hand, end-to-end VLA policies prove insufficient for scenarios demanding explicit physical modeling, as they fail to generalize to non-obvious maneuvers like maintaining steady force (T4). On the other hand, the L2R baseline demonstrates the fragility of static code generation; it fails tasks like the wall-flip (T6: 1/10) because it generates fixed cost functions without explicitly formulating a contact strategy or employing an online refinement mechanism to handle physical deviations. In contrast, our framework excels by combining a dedicated contact strategy with an adaptive LLM that directly formulates and iteratively refines cost functions, enabling robust zero-shot execution across all dynamic interaction regimes.

\subsubsection{Comparison to Human-Designed Cost Functions}
The two human baselines approximate an upper bound from carefully engineered, task-specific objectives. As expected, the \emph{Expert (FSM)} variant achieves the strongest overall performance, and the single-stage expert design remains competitive, particularly on simpler tasks such as T2–T4, where CoRAL largely matches but does not surpass its success rate (Table~\ref{tab:comprehensive_results}). On more sequential and contact-heavy tasks (T1, T5, T6), CoRAL narrows the gap to the expert, achieving higher success rates than the single-stage baseline while remaining below the FSM upper bound. This shows that our LLM-based controller can recover much of the structure of expert-designed costs automatically, substantially reducing manual tuning effort while approaching expert-level performance on the hardest tasks.

\subsubsection{Ablation Study Analysis (RQ2)}
Our ablation studies clearly demonstrate the necessity of each component in our architecture.

\textbf{The Synergy of Separated VLM/LLM Roles:} The \textit{CoRAL (Unified VLM)} variant, which tasked a single VLM with both perception and planning, failed on nearly all complex tasks. This starkly illustrates our core hypothesis: separating the role of a VLM for perception from a dedicated LLM for strategy formulation is crucial for robust performance. The specialized modules provide more reliable and structured outputs for the planner.

\textbf{The Importance of Online Refinement:} The \textit{w/o Refinement} variant showed a dramatic performance drop in multi-stage tasks like T1 (Push and Pick Board), with success falling from 5/10 to 0/10. In this task, the initial plan often failed because the VLM's initial friction estimate was slightly off, causing the board to slip during the pick. The full CoRAL framework, however, used the outer loop for the LLM to diagnose this from the physical outcome, refine the friction parameter in its world model, and successfully complete the task. This shows the system's ability to learn from failure.

\textbf{The Benefit of Experience Reuse:} The full framework \textit{with Memory} consistently achieved the highest success rates. For instance, in T1 and T3, memory boosted the success rate from 2/10 to 5/10 and 9/10 to 10/10, respectively. By retrieving a successful ``push-to-edge" strategy from a past experience, the system provided the planner with a superior initialization, accelerating convergence and leading to more robust solutions.

\textbf{The Criticality of a Dedicated Pose Estimator:} The \textit{w/o Pose Tracking} ablation, which removed FoundationPose and relied solely on the VLM for pose estimation, resulted in a catastrophic failure across all tasks (0/10 success). The VLM, while powerful for semantic understanding, is ill-suited for the precision required by 6-DoF pose tracking through dynamic interactions. It frequently produced trivial or physically impossible pose estimations (``hallucinations") that rendered the planner's output useless. This result provides conclusive evidence that a dedicated, high-fidelity pose estimator is not merely beneficial but an essential component of our architecture, serving as the geometric foundation upon which all subsequent physical reasoning is built.

\subsubsection{Robustness Analysis (RQ3)}

\textbf{Analysis of LLM-Guided Contact Strategy:}
To isolate the contribution of the LLM's initial contact strategy ($C_0$), we conducted a targeted ablation on the challenging ``Flip with Wall" task (T6). We compared the performance of our full framework against a variant where the LLM only provided the cost function ($J_0$), forcing the MPPI planner to discover useful contact points through its own sampling mechanism. Crucially, this variant effectively mimics the operational capability of standard foundation-model-based planner baseline like L2R~\cite{yu2023language}, which rely solely on the optimizer to discover contact modes.

\begin{figure*}[t]
\centering
\includegraphics[width=0.8\linewidth]{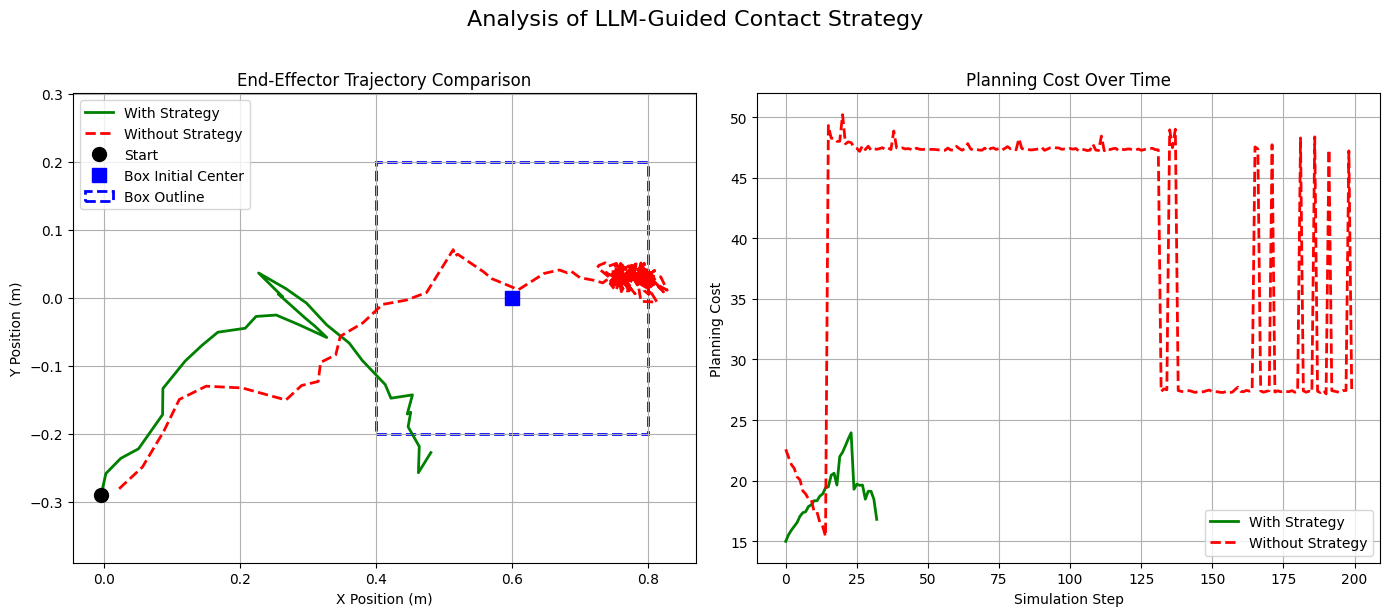}
\caption{
Ablation of the LLM-guided contact strategy on the ``Flip with Wall" task. 
\textbf{(Left)} The trajectory with the LLM's strategy (green) is direct 
and efficient, while the unguided trajectory (red) is erratic. 
\textbf{(Right)} The planning cost for the guided agent is significantly 
lower and more stable, indicating an easier optimization problem.
}
\label{fig:contact_strategy_ablation}
\end{figure*}

The guided trajectory (With Strategy, green) is direct and purposeful, immediately moving the end-effector to the correct corner of the box to initiate the flip  (Fig.~\ref{fig:contact_strategy_ablation}). In contrast, the unguided trajectory (Without Strategy, red) is chaotic and inefficient, exploring large, irrelevant portions of the workspace. The planning cost for the unguided agent remains high and erratic, indicating a constant struggle to find a viable plan. This difference is confirmed by the quantitative results: the guided approach was \textbf{83.9$\%$ more efficient} requiring fewer control steps (32 vs.~199 steps) and the end-effector traveled a \textbf{63.9$\%$ shorter path} (1.33m vs.~3.69m). This analysis provides clear evidence that the LLM's symbolic contact strategy is critical for transforming a computationally intractable, long-horizon contact problem into a solvable one by intelligently pruning the vast action search space.

\textbf{Robustness of Online Parameter Adaptation:}
Beyond strategy and cost function refinement, CoRAL's `Online Adaptation' module, driven by the LLM, exhibits a crucial ability to correct the agent's internal world model online. To demonstrate this, we ran the same experiment both in simulation and on the real system to verify that the adaptation mechanism remains effective under real-world unmodeled effects and sensing/actuation noise, intentionally initializing the \textit{Planning World} with a severely overestimated mass (2.0kg vs.~a ground truth of 0.25kg) and friction coefficient (0.9 vs.~0.5) for the cutting board. These initial biases represent a severe sim-to-real gap or a VLM hallucination.

\begin{figure}[t]
    \centering
    \includegraphics[width=\linewidth]{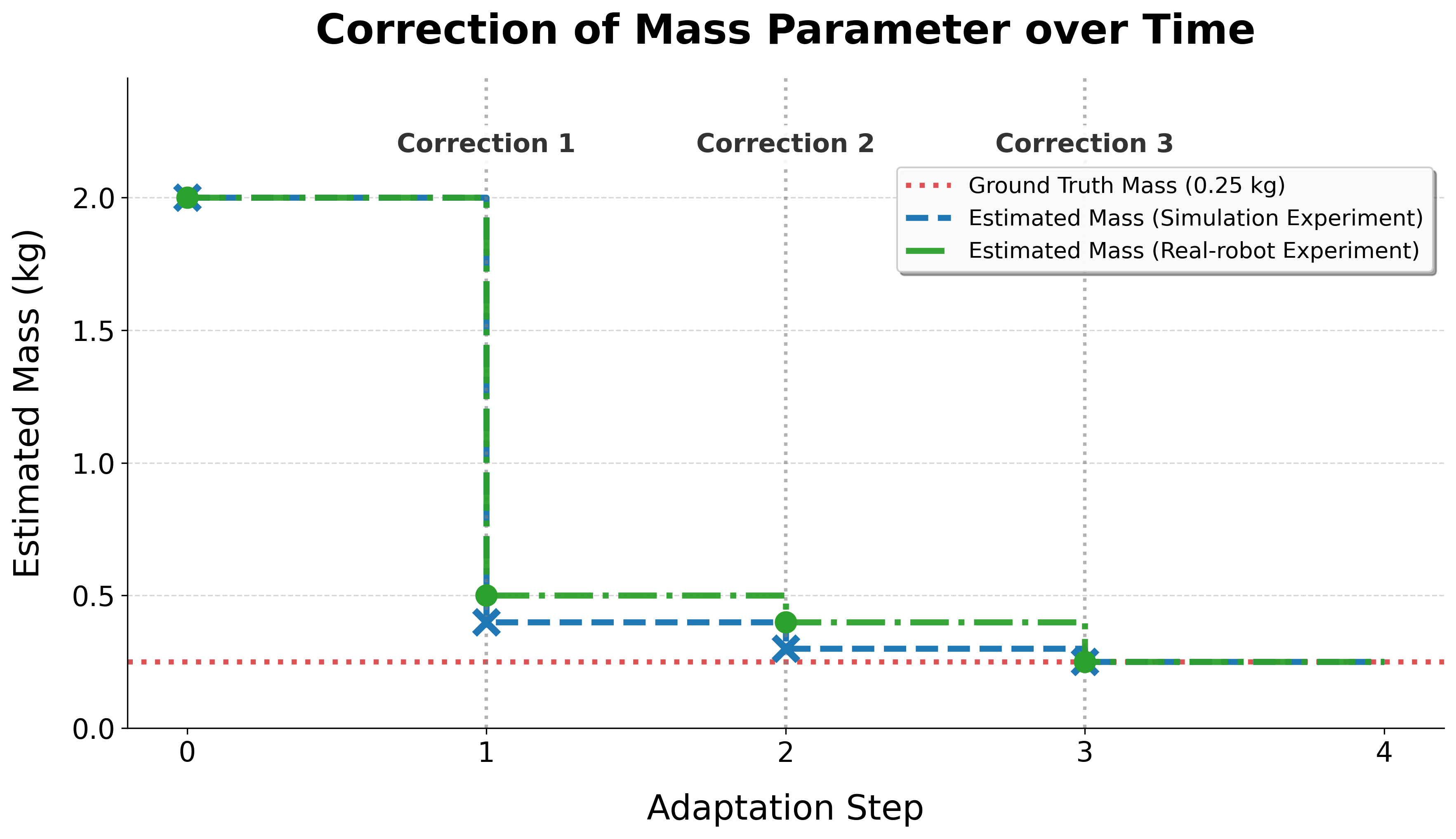}
    \caption{Online parameter (mass) adaptation in simulation and real-world experiments.}
    \label{fig:parameter_adaptation}
    \vspace{-15pt}
\end{figure}

Figure~\ref{fig:parameter_adaptation} shows the adaptation process. Given the execution history, the LLM's `Online Adaptation' module identified that the board was not moving as expected despite high pushing force and updated the physical parameters. Through an iterative refinement process, it progressively adjusted its estimated mass and friction parameters. After several adaptation cycles, the agent's belief about both mass and friction converged remarkably close to their true values. This online correction of physical parameters is fundamental to the framework's robustness, allowing it to overcome initial environmental mischaracterizations and successfully execute contact-rich tasks that would otherwise fail due to a misaligned internal world model.

\textbf{Sensitivity to Pose and Physical-Parameter Estimates:}
To further quantify the robustness of CoRAL to imperfect world-model initialization, we performed a sensitivity analysis on the ``Push and Pick Cutting Board'' task (T1). 
For physical parameter sensitivity, the cutting-board mass was uniformly sampled from $[0.4,0.8]$kg, and the sliding friction coefficient from $[0.3,0.6]$. 
Across 100 samples, the VLM-only estimates achieved a mean absolute error (MAE) of $0.29$kg for mass and $0.14$ for friction. 
These results support our design choice of treating VLM outputs as coarse semantic priors rather than accurate physical measurements. 
After four online refinement cycles, the errors decreased to $0.11$kg and $0.06$, respectively, improving the task success rate from $2/10$ to $5/10$.

For the same T1 task, Table~\ref{tab:sensitivity_results} summarizes the effect of replacing estimated quantities with oracle information. 
Using ground-truth physical parameters alone did not further improve performance beyond the refined setting, suggesting that CoRAL can largely compensate for moderate mass and friction errors through replanning and online adaptation. 
In contrast, replacing the estimated object poses with ground-truth poses increased the success rate to $7/10$, and using both ground-truth poses and ground-truth physical parameters further improved it to $8/10$. 
This indicates that pose-estimation errors, especially during occlusions caused by direct end-effector contact, are the dominant bottleneck in this task. 
In our cutting-board trials, FoundationPose achieved a mean ADD error of $13.2$mm over 10 episodes, while $88.2\%$ of frames had ADD below $5$mm. 
The larger mean error is mainly caused by short tracking failures during contact-rich phases. 
Overall, physical-parameter errors primarily affect the magnitude of predicted object motion and can often be corrected through feedback and replanning, whereas pose failures directly corrupt the replanning state and are therefore more detrimental.

\begin{table}[t]
\centering
\caption{Cutting-board success rates under different pose and physical-parameter settings.}
\label{tab:sensitivity_results}
\scriptsize
\begin{tabular}{p{0.68\linewidth} c}
\toprule
\textbf{Condition} & \textbf{Success Rate} \\
\midrule
VLM-estimated parameters, no parameter refinement & $2/10$ \\
VLM-estimated parameters, with parameter refinement & $5/10$ \\
Ground-truth physical parameters & $5/10$ \\
Ground-truth poses & $7/10$ \\
Ground-truth poses + ground-truth physical parameters & $8/10$ \\
\bottomrule
\end{tabular}
\vspace{-8pt}
\end{table}

\textbf{Sequential Reasoning and Experience Reuse in the Cutting Board Task:} The ``Push and Pick Cutting Board" task (T1) tests long-horizon sequential manipulation, requiring a stable push to expose the board followed by a precise grasp. As evidenced by Table~\ref{tab:comprehensive_results}, this challenge highlights the importance of two core components: online adaptation and experience reuse.

First, long-horizon tasks are sensitive to model errors that accumulate over time. This is demonstrated by the \textit{w/o Refinement} ablation, which failed entirely (0/10 success rate), mirroring the static L2R baseline~\cite{yu2023language}. While the initial plan was often sufficient for the push, slight inaccuracies in estimated friction caused the board to end in an unexpected pose, leading to a failed grasp. Our full model, however, leverages the outer loop to learn from the push outcome, allowing the `LLM (Online Adaptation)' to refine friction estimates and update the plan for the subsequent pick.

Second, this task illustrates the benefit of the `Memory Unit'. Including the memory module boosted the success rate from 2/10 to 5/10. This indicates that after a single completion, the system stores the successful interaction context (refined parameters and strategy) to provide a superior initialization for future attempts. This demonstrates that CoRAL can reuse prior successful experience, highlighting a clear path towards few-shot performance improvements as it gathers successful episodes.

\textbf{Explainability and Automated Failure Recovery}
A key advantage of our neuro-symbolic design is its inherent explainability, particularly during failure recovery. Unlike opaque end-to-end models, CoRAL can articulate \textit{why} it failed and \textit{what} it is doing to correct its plan. We demonstrate this with a scenario where the ``Flip with Wall" task persistently fails, triggering the Outer Loop.

Instead of just outputting a new set of parameters, the `LLM (Online Adaptation)' module provides a full natural language diagnosis of the failure and a detailed log of the corrective actions it is taking. The LLM provided a correct natural language diagnosis of a poorly weighted cost function and proceeded to adjust the specific weights to remedy the failure (see Supplementary Material).

\textbf{LLM Failure Modes:}
While CoRAL avoids using the LLM as a direct controller, failures can still arise from the symbolic objectives it generates. 
We observed two main failure modes. 
First, the LLM occasionally produced highly imbalanced cost weights, e.g., $w_d : w_c = 1000:1$, causing MPPI to effectively ignore some objectives. 
We mitigated this by prompting the LLM to justify the relative importance of each term and keep weights within comparable numerical ranges, typically $[0.1,10]$. 
Second, in multi-stage tasks such as T1 (\emph{Push and Pick Cutting Board}) and T6 (\emph{Flip with Wall}), the LLM sometimes produced ambiguous or incomplete stage-transition conditions, leading to premature or delayed phase switches; representative examples are provided in the Supplementary Material.
These structural errors occurred in fewer than $10\%$ of trials but were more consequential than small weight errors. 
Importantly, CoRAL's modular design makes such failures interpretable: they appear as persistently high rollout costs or mismatches between the intended subgoal and observed state evolution, allowing the outer-loop refinement module to update the cost structure or stage logic.

\subsubsection{Real-World Validation (RQ4)}
\label{sec:real_world}

Finally, to validate the reliability of CoRAL under physical constraints that cannot be perfectly simulated (e.g., sensor noise, unmodeled friction, and calibration errors), we deployed the system zero-shot on the physical Franka Emika Panda robot. We evaluated the full suite of six tasks without any real-world fine-tuning (Fig.~\ref{fig:real_robot}).

\begin{table}[h]
\centering
\caption{\textbf{Real-World Experimental Results.} CoRAL is evaluated on the physical robot across all six tasks without real-world fine-tuning.}
\label{tab:real_world_results}
\resizebox{0.95\linewidth}{!}{%
\begin{tabular}{l c c}
\toprule
\textbf{Task} & \textbf{Success Rate} & \textbf{Avg. Exec. Time (s)} \\
\midrule
T1: Push and Pick Cutting Board & 4/10 & 21.6 $\pm$ 5.5 \\
T2: Pick Box & 10/10 & 16.7 $\pm$ 1.1 \\
T3: Pick and Place in Clutter & 10/10 & 22.0 $\pm$ 1.5 \\
T4: Push with Constant Force & 9/10 & 11.7 $\pm$ 1.9 \\
T5: Flip Box & 7/10 & 9.2 $\pm$ 2.6 \\
T6: Flip with Wall & 6/10 & 25.3 $\pm$ 4.9 \\
\bottomrule
\end{tabular}%
}
\end{table}

\textbf{Sim-to-Real Robustness:} As shown in Table~\ref{tab:real_world_results}, CoRAL demonstrates strong sim-to-real transfer capabilities. While standard tasks (T2, T3) achieved perfect success rates, the framework also maintained robust performance on contact-critical tasks (T1, T4, T5). A notable observation was in the ``Push and Pick" task (T1), where real-world surface friction varied significantly from the simulation. In these cases, the online adaptation loop (discussed in Section~\ref{sec:experiments}) successfully diagnosed the slippage and updated the friction parameters in real-time to salvage the trial.

\begin{figure}[t]
\begin{center}
\includegraphics[width=\linewidth]{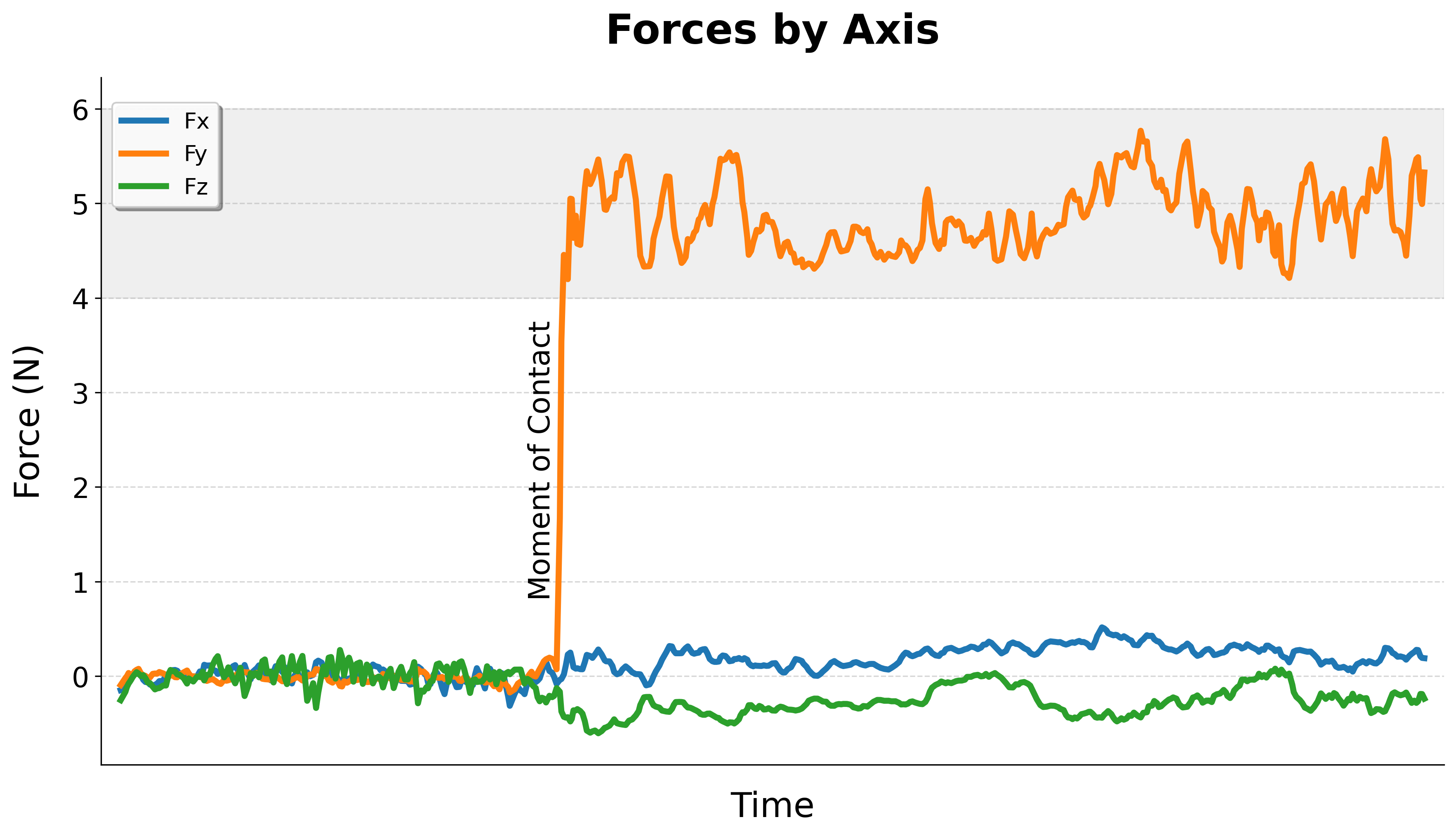} %
\end{center}
\caption{\textbf{Real-World Force Regulation Profile (T4)} The plot shows the measured end-effector force over time during the ``Push with Constant Force" task where the intended push direction is $+y$. The shaded region represents the target force range the LLM is instructed with. CoRAL successfully modulates the control actions to maintain contact forces within the desired bounds.} 
\label{fig:force_profile}
\vspace{-15pt}
\end{figure}

\textbf{Force Regulation:} The system's ability to handle active force constraints is highlighted in Task 4. Figure~\ref{fig:force_profile} shows the force profile from a real-world trial. Despite the inherent noise in the physical force/torque sensor, the MPPI controller—guided by the LLM's cost function—effectively regulated the interaction force within the target bounds (approx. 5N). This confirms that CoRAL's decoupled architecture not only plans for geometry but effectively closes the loop on force dynamics, bridging the gap between high-level semantic planning and low-level compliance.

\section{Limitations \& Conclusion}
\label{sec:conclusion}

In this paper, we introduced \textbf{CoRAL}, a novel framework that addresses the challenges of zero-shot, contact-rich manipulation. Our approach departs from conventional end-to-end paradigms by integrating foundation models with a reactive controller. Experiments on challenging tasks demonstrate that this modular, synergistic design enables the system to adapt to unseen scenarios without prior demonstrations, significantly enhancing both performance and explainability over monolithic approaches. While promising, the framework's performance is currently contingent on the fidelity of the vision-based world model and is subject to latency constrained by foundation model inference. 
These limitations and future research directions are discussed in detail in the supplementary material. We believe this hybrid paradigm—coupling large-scale, pre-trained knowledge with rigorous real-time control—is a promising direction for creating more capable and physically intelligent robotic agents.

\bibliographystyle{plainnat}
\bibliography{cit}
\clearpage
\onecolumn
\appendices
\setcounter{section}{0}
\setcounter{figure}{0}
\setcounter{table}{0}
\setcounter{equation}{0}
\renewcommand{\thefigure}{S\arabic{figure}}
\renewcommand{\thetable}{S\arabic{table}}
\renewcommand{\thesection}{S\arabic{section}}
\renewcommand{\theequation}{S\arabic{equation}}

\section{Preliminaries}
\label{sec:preliminaries}

In this section, we introduce the Model Predictive Path Integral (MPPI) controller, which forms the core of our methodology, and provide a formal problem formulation for the contact-rich manipulation tasks we address.

\subsection{Model Predictive Path Integral (MPPI)}

Model Predictive Path Integral (MPPI)~\cite{williams2017model} is a sampling-based Model Predictive Control (MPC)~\cite{garcia1989model} algorithm designed to solve stochastic optimal control problems. It is particularly effective for systems with nonlinear and complex dynamics. MPPI operates by simulating thousands of potential control sequences in parallel from the current state to determine the optimal subsequent control input.

Consider a system with discrete-time stochastic dynamics described by $ \mathbf{x}_{t+1} = f(\mathbf{x}_t, \mathbf{u}_t) + \boldsymbol{\epsilon}_t $, where $ \mathbf{x}_t $ is the state of the system, $ \mathbf{u}_t $ is the control input, and $ \boldsymbol{\epsilon}_t $ represents system noise. The objective of MPPI is to find a control sequence $ \mathbf{U} = \{\mathbf{u}_0, \mathbf{u}_1, \dots, \mathbf{u}_{H-1}\} $ that minimizes an expected cost function:
\begin{equation}
J(\mathbf{U}) = \mathbb{E}\left[\phi(\mathbf{x}_H) + \sum_{t=0}^{H-1} q(\mathbf{x}_t, \mathbf{u}_t) \right]
\end{equation}
Here, $ \phi(\mathbf{x}_H) $ is the terminal cost, and $ q(\mathbf{x}_t, \mathbf{u}_t) $ is the running (or stage) cost, which typically includes terms for tracking error and control effort. $ H $ is the planning horizon.

At each time step, MPPI samples $ K $ candidate control sequences by adding random noise perturbations to a nominal sequence. These are rolled out in simulation to get trajectories, and the total cost $ S_k $ for each is computed. The optimal control is then calculated via an exponential weighting of these costs. This process is repeated at each time step following the receding horizon principle. In our approach, the symbolic reasoning provided by the LLM forms the initial structure for the running cost function $ q(\cdot) $.

 \subsection{MPPI with Adaptive Temperature (\texorpdfstring{$\lambda$}{lambda}) via ESS}
\label{sec:mppi_auto_lambda}

The key hyperparameter for the controller is the temperature $\lambda$. Smaller $\lambda$ concentrates probability mass on the best trajectories (more exploitative updates), whereas larger $\lambda$ yields more uniform weights (more exploratory but weaker updates). In practice, a fixed $\lambda$ can be brittle because the value of $\lambda$ is tied to the magnitude of the cost function which is solely determined by the LLM generated weights and functionals. We therefore select $\lambda$ adaptively at every planning iteration.

At planning iteration $n$, we maintain a nominal horizon-$H$ control sequence
\begin{equation}
U^{(n)}=\big(U^{(n)}_0,\dots,U^{(n)}_{H-1}\big), \qquad U^{(n)}_t\in\mathbb{R}^d,
\end{equation}
and sample $K$ perturbed candidates. With state cost $c:\mathcal{X}\to\mathbb{R}$, each sample $i$ is formed as
\begin{equation}
u_{i,t}=\mathrm{clip}\!\left(U^{(n)}_t+\varepsilon_{i,t},-u_{\max},u_{\max}\right),
\end{equation}
rolled out from the current state $x^{(n)}$, and scored by the average horizon cost
\begin{equation}
J_i=\frac{1}{H}\sum_{t=0}^{H-1} c(x_{i,t+1}).
\end{equation}
For numerical stability we shift costs by the best sample:
\begin{equation}
S_i = J_i-\min_j J_j.
\end{equation}

Given a temperature $\lambda>0$, MPPI assigns exponential weights
\begin{equation}
w_i(\lambda)=\exp\!\left(-\frac{S_i}{\lambda}\right),\qquad
W_i(\lambda)=\frac{w_i(\lambda)}{\sum_{j=1}^K w_j(\lambda)}.
\end{equation}

We select $\lambda$ online using an effective-sample-size (ESS) target. We define
\begin{equation}
\mathrm{ESS}(\lambda)=\frac{1}{\sum_{i=1}^K W_i(\lambda)^2}\in[1,K],\qquad
E_{\text{tgt}}=\phi K,
\end{equation}
with target fraction $\phi\in(0,1]$. We choose $\lambda^{(n)}$ via bisection to satisfy $\mathrm{ESS}(\lambda)\approx E_{\text{tgt}}$. We initialize
\begin{equation}
\lambda_{\mathrm{lo}}=10^{-8},\qquad
\lambda_{\mathrm{hi}}=5\cdot \max\!\big(10^{-3},\max_i S_i\big),
\end{equation}
and iterate $T=25$ steps:
\begin{equation}
\lambda=\tfrac{1}{2}(\lambda_{\mathrm{lo}}+\lambda_{\mathrm{hi}}),\quad
\text{if } \mathrm{ESS}(\lambda)<E_{\text{tgt}} \Rightarrow \lambda_{\mathrm{lo}}\leftarrow\lambda,\;
\text{else } \lambda_{\mathrm{hi}}\leftarrow\lambda.
\end{equation}
We output $\lambda^{(n)}=\tfrac{1}{2}(\lambda_{\mathrm{lo}}+\lambda_{\mathrm{hi}})$.

Using $\lambda^{(n)}$, we update the nominal controls by the weighted perturbation average:
\begin{equation}
\Delta U_t=\sum_{i=1}^K W_i(\lambda^{(n)})\,\varepsilon_{i,t},\qquad
U^{(n+1)}_t=\mathrm{clip}\!\left(U^{(n)}_t+\beta\,\Delta U_t,-u_{\max},u_{\max}\right).
\end{equation}

\subsection{Problem Formulation}

In this work, we address long-horizon, contact-rich manipulation tasks specified by visual and linguistic commands. Our goal is to develop a robotic system capable of interacting with objects of unknown physical properties (e.g., mass, friction) and generalizing to new scenarios in a zero-shot manner.

We formulate the problem as a POMDP. The state of the system, $ \mathbf{x}_t \in \mathcal{X} $, includes the robot's state $ \mathbf{x}_r(t) $ and the states of environmental objects $ \mathbf{x}_o(t) $. The action space, $ \mathcal{U} $, consists of continuous control commands applicable to the robot's end-effector.

At the beginning of each task, the system receives an RGB-D image $I$, a natural language instruction $T$, and the corresponding 3D object models $M$. The system's objective is not to learn a fixed policy, but rather to compute, at each step, a control action $u_t$ that leads to a sequence of actions $ \mathbf{U} = \{\mathbf{u}_0, \dots, \mathbf{u}_{H-1}\} $ that successfully completes the task.

The core challenge is to bridge the gap from high-level, multimodal inputs ($I, M, T$) to these low-level, continuous control actions. Our approach decomposes this problem into two stages:
\begin{enumerate}
    \item \textbf{Strategy Formulation:} We use the vision module and LLM to translate the multimodal inputs ($I, M, T$) into an initial cost function $ J_0 $ and contact candidates $ C_0 $ for the planner.
    \item \textbf{Online Planning and Control:} We employ the MPPI planner to compute the optimal action sequence online, guided by the LLM's strategy and adapted using real-time sensory feedback.
\end{enumerate}

This formulation accurately frames our system as an online planner that reasons and computes actions on the fly, rather than an agent executing a pre-learned, static policy.

\section{End-to-End Workflow: A Visual Overview}
\label{app:end_to_end_workflow}

\begin{figure*}[t!]
    \centering
    \includegraphics[width=\textwidth]{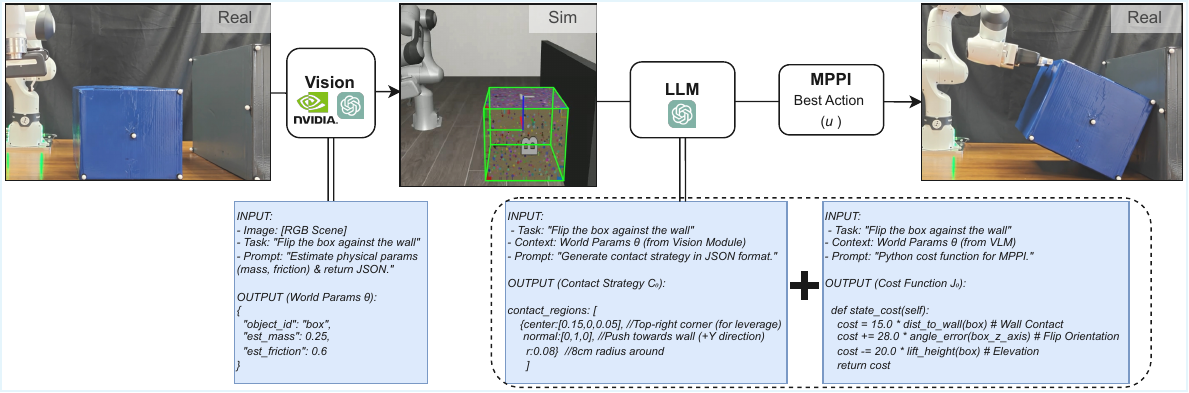}
    \caption{
    \textbf{End-to-end workflow execution trace.} This figure illustrates the data flow using representative snippets of the actual inputs and outputs exchanged between modules, complementing the architectural overview in Figure~2. 
    \textbf{Left (Internal World Model):} The vision module tracks 6-DoF object poses via FoundationPose~\cite{wen2024foundationpose} (in simulation) or motion capture (in real-world), and queries the VLM to estimate physical parameters ($\theta$: mass, friction) in JSON format to initialize the planning world. 
    \textbf{Center (Task Formulation):} The LLM generates two structured outputs from the task description: (1) semantic contact region descriptors (center, normal, radius), and (2) executable Python cost functions that define the MPPI objectives. 
    \textbf{Right (Action \& Feedback):} MPPI executes at 10Hz with reactive control, while the outer LLM loop asynchronously refines both world model parameters and the strategic cost structure based on interaction feedback. 
    Full prompts and implementation details are provided in Sections~\ref{app:vlm_prompt}, \ref{app:cost_generation_prompts}, and \ref{app:contact_strategy_generation}.
    }
    \label{fig:end_to_end_workflow}
\end{figure*}

Before diving into experimental details and implementation specifics, we provide a concrete illustration of CoRAL's complete pipeline in Figure~\ref{fig:end_to_end_workflow}. This figure complements the architectural diagram in Figure~2 of the main paper by showing the \textbf{actual data exchanged} between modules: VLM-estimated parameters in JSON, LLM-generated Python cost functions, and the refinement feedback loop.

The workflow demonstrates how CoRAL bridges semantic reasoning with physical control:
\begin{enumerate}
    \item \textbf{Internal World Model:} Vision module uses FoundationPose~\cite{wen2024foundationpose} for continuous 6-DoF object tracking (replaced by motion capture in real-world experiments), then queries the VLM to estimate physical parameters ($\theta$) in JSON format for initializing the simulated planning world.
\item \textbf{Task Formulation:} LLM translates natural language into two structured outputs: (a) executable Python \textbf{cost functions} that define MPPI objectives, and (b) \textbf{contact region descriptors} (JSON format: center, normal, radius) that guide a geometric sampler to generate candidate contact points from semantically meaningful surface areas, dramatically narrowing the search space.
    \item \textbf{Action \& Feedback:} High-frequency control (MPPI) + low-frequency adaptation (LLM diagnostics).
\end{enumerate}

Sections~\ref{app:vlm_prompt}--\ref{app:contact_strategy_generation} provide the detailed prompts, code, and JSON schemas illustrated here.

\section{Experimental Details}
\label{sec:experimental details}

 The VLM and LLM modules were implemented using the GPT-4o API. The MPPI controller was integrated on top of the \textsc{robosuite}/\textsc{MuJoCo} environment. To improve computational efficiency, our MPPI implementation parallelizes the rollout of $K=256$ trajectories over a planning horizon of $H=32$ steps at each control cycle on the CPU. The selection $\lambda$ is detailed in the previous section.  The target fraction, $\phi$, is set to $0.2$ for all experiments
 
The entire framework runs on a desktop PC equipped with an Intel Core i9-13900K CPU, 64\,GB RAM, and a single NVIDIA RTX 4060 Ti GPU.

For real-world experiments, we use a Franka Emika Panda robot with a parallel-jaw gripper. Object poses are tracked using a six-camera Vicon Vero motion-capture setup.

\section{Qualitative Comparison with State-of-the-Art VLA Manipulation Frameworks}

\begin{table*}[h]
\centering
\caption{Comparison with State-of-the-Art VLA Manipulation Frameworks}
\label{tab:related_work_comparison_expanded}
\resizebox{\textwidth}{!}{%
\begin{tabular}{l cccc}
\toprule
\textbf{Framework} & \textbf{Primary Modality} & \textbf{Planning \& Control Strategy} & \textbf{Reasoning Mechanism} & \textbf{Data Requirement} \\
\midrule
OpenVLA~\cite{kim2025openvla} & Vision, Language & End-to-End Learned Policy (Action Token Prediction) & Implicit (in VLM backbone) & Large-scale Imitation Learning Demos \\
$\pi_{0.5}$~\cite{intelligence2504pi0} & Vision, Language & End-to-End Learned Policy (Flow Matching) & Implicit (in VLM backbone) & Large-scale Imitation Learning Demos \\
\midrule
ForceVLA~\cite{yu2025forcevla}, TLA~\cite{hao2025tla} & Vision, Language, \textbf{Tactile/Force} & End-to-End Learned Policy & Implicit (in network weights) & Large-scale Tactile/Force Demos \\
VLA-Touch~\cite{bi2025vlatouch} & Vision, Language, \textbf{Tactile} & VLA Policy + Tactile-based Refinement Controller & Explicit VLM Planning + Semantic Tactile Feedback & Leverages pretrained models; no VLA fine-tuning \\
\midrule
ThinkAct~\cite{huang2025thinkact}, ECoT~\cite{zawalski2024ecot} & Vision, Language & End-to-End Learned Policy & \textbf{Explicit LLM Reasoning} (Chain-of-Thought) & Large-scale Imitation Learning Demos \\
OneTwoVLA~\cite{lin2025onetwovla} & Vision, Language & Unified Policy (Adaptive Acting \& Reasoning) & \textbf{Explicit LLM Reasoning} (System 2) & Imitation Demos + Synthetic Reasoning Data \\
MolmoAct~\cite{lee2025molmoact} & Vision, Language & Multi-stage Pipeline (Perception, Spatial Plan, Action) & \textbf{Explicit Spatial Reasoning} (Trajectory Traces) & Large-scale Imitation Learning Demos \\
\midrule
IMPACT~\cite{ling2025impact} & Vision, Language & VLM-based Static Cost Map + RRT* & Implicit (semantic object labeling) & N/A (Planner-based) \\
VLMPC~\cite{zhao2024vlmpc} & Vision, Language & VLM-guided Model Predictive Control (MPC) & \textbf{Explicit VLM Reasoning} (for cost \& sampling) & N/A (Planner-based) \\
L2R~\cite{yu2023language} & Language & LLM-generated Reward Function + MPC & \textbf{Explicit LLM Reasoning} (Static Code Gen.) & N/A (Planner-based) \\
\midrule
\textbf{CoRAL (Ours)} & Vision, Language, \textbf{Tactile/Force} & \textbf{LLM-guided MPPI + Reactive Control} & \textbf{Explicit LLM Reasoning} (Strategy + Adaptation) & \textbf{Zero-Shot (No Demos)} \\
\bottomrule
\end{tabular}%
}
\end{table*}

As the comparative analysis in Table~\ref{tab:related_work_comparison_expanded} illustrates, the field of robotic manipulation has historically involved a trade-off. End-to-end VLA models achieve impressive behaviors but are fundamentally constrained by large-scale demonstration datasets, while traditional planner-based systems are zero-shot but often lack high-level semantic reasoning. CoRAL is designed to synthesize the strengths of these disparate paradigms. To the best of our knowledge, CoRAL is the first framework to simultaneously integrate \textbf{explicit LLM-driven strategy formulation} with a \textbf{dynamic, reactive controller} that leverages real-time \textbf{tactile and force feedback}, all while operating in a \textbf{zero-shot} manner that completely eliminates the need for prior demonstration data.

\section{Discussion \& Limitations}
\label{sec:discussion}

Our work introduces CoRAL, a framework that represents a deliberate architectural shift away from the prevailing end-to-end paradigm for Vision-Language-Action (VLA) models. By decoupling high-level reasoning from low-level, physics-based planning and control, we address several fundamental challenges in contact-rich manipulation, particularly regarding explainability, data efficiency, and physical grounding.

\subsection{Discussion}

Our experimental results validate the core hypotheses of CoRAL, demonstrating that a modular, neuro-symbolic architecture can overcome the fundamental limitations of end-to-end models in complex, contact-rich manipulation. We discuss the key implications of our findings below.

\textbf{The Synergy of Grounded Reasoning and Reactive Control:}
Our experiments reveal a clear synergy between high-level reasoning and low-level reactive control. The performance gap between CoRAL and end-to-end baselines like OpenVLA, especially in tasks requiring non-trivial strategies (e.g., T6: Flip with Wall), highlights the brittleness of purely imitative policies. These models fail because their training data lacks examples of using the environment as a tool. In contrast, CoRAL's success stems from its ability to reason: the LLM formulates an explicit optimization problem ($J_0$) that defines the task's success conditions, while the MPPI controller finds a physically plausible solution. This makes the system's intent transparent and grounds abstract reasoning in a formal control framework, a more robust approach than conditioning a black-box policy as is done in works like ECoT~\cite{zawalski2024ecot}.

\textbf{Online Adaptation as an Alternative to Large-Scale Tactile Datasets:}
A significant implication of our work is a path away from the data-hungry paradigm of modern robotics. State-of-the-art methods like ForceVLA~\cite{yu2025forcevla} achieve impressive results by incorporating tactile feedback, but this requires creating massive, specialized demonstration datasets. Our results demonstrate a more data-efficient alternative. CoRAL also uses real-time force feedback, but its role is redefined: it serves as a signal for \textit{online adaptation}, not offline imitation. The success of our outer feedback loop, particularly in tasks where initial parameter estimates were deliberately inaccurate, proves that the LLM can diagnose physical failures and refine its world model on the fly. This ability to learn from direct interaction significantly lowers the barrier to entry for creating sophisticated, contact-aware robots without relying on pre-collected, large-scale tactile data.

\textbf{Zero-Shot Planning and the Path to Lifelong Learning:}
Perhaps the most significant result is CoRAL's ability to perform zero-shot planning for novel tasks. This capability stems from its core design: instead of learning \textit{how to act}, it leverages the pre-existing knowledge of foundation models to reason about \textit{how to plan}. By generating a cost function from a single image and a language command, the system dynamically tackles new objectives. The consistent performance boost provided by the `Memory Unit' in our ablation studies further points towards a lifelong learning capability. By retrieving and bootstrapping successful strategies, the system becomes more efficient and robust over time, a contrast to the static nature of policies that require extensive fine-tuning or retraining to adapt~\cite{kim2025openvla,lee2025molmoact}.

\subsection{Limitations and Future Work}
\label{sec:limitations}

Despite its promising results, CoRAL has several limitations that define clear directions for future research.

\textbf{Fidelity of the Internal World Model:} The entire framework is predicated on the quality of the simulated planning world constructed by the vision module. This dependency is a significant limitation; the performance of the MPPI planner is directly correlated with how accurately this internal model reflects real-world physics. We can think of this simulated planning environment as the robot's ``mind," where it mentally rehearses actions before execution. The better this mental model, the more seamlessly the robot can translate its plans into successful evaluation-world actions. Currently, the system is vulnerable to inaccuracies in object pose estimation, and VLM ``hallucinations" or inaccuracies---misjudging an object's material could lead to a grossly incorrect estimate for mass or friction. While our feedback loop is designed to correct for such errors, a sufficiently poor initialization could prevent the planner from converging. Future work should focus on creating higher-fidelity internal world model, potentially by learning residual dynamics models online to capture unmodeled effects (e.g., non-rigid dynamics, complex friction) and better bridge the sim-to-real gap.

\textbf{Reliance on Generalist LLMs for Strategy Formulation:} A core component of our system is the LLM's ability to generate a viable cost function for the MPPI planner. We currently use a generalist, off-the-shelf LLM (GPT-4o), which performs remarkably well but is not specialized for robotics or physics-based planning. The quality and coherence of the generated cost function are not guaranteed for entirely novel or abstract tasks that fall outside the LLM's vast but general pre-training data. A promising direction for future work is to fine-tune or develop LLMs specifically for the task of generating optimization objectives for robotic control, potentially leading to more robust and efficient strategy formulation.

\section{Internal World Model for Planning and Adaptation}
\label{app:world_model}

The core of CoRAL's reasoning capability lies in its use of an explicit internal world model. This model is a simulated, physics-aware representation of the real world, constructed and parameterized by the vision module's output, $\theta$. This internal model is not merely a passive environment for the planner; it serves two critical functions in our architecture:

\begin{enumerate}
    \item \textbf{A Sandbox for Mental Rehearsal:} The MPPI controller leverages this world model to perform a number of parallel ``mental rehearsals" (rollouts) of potential action sequences. This allows the agent to anticipate the physical consequences of its actions—such as contact forces and object displacement—before executing them in the evaluation world, enabling proactive and intelligent decision-making.

    \item \textbf{An Adaptable Belief State:} This world model represents the agent's current ``belief" about the physical properties of its environment. Crucially, this belief is not static. The outer feedback loop, driven by the `LLM (Online Adaptation)' module, directly refines this model by updating its parameters ($\theta$) based on discrepancies between predicted and observed outcomes. This turns the world model into a dynamic, adaptable belief state that is continuously improved through physical interaction, connecting our work to principles of online system identification and model-based reinforcement learning.
\end{enumerate}

Below is the direct natural language output from the `LLM (Online Adaptation)' module after being triggered by persistent failures in the ``Flip with Wall" task. The LLM was provided with the episode history ($E_t$) and prompted to diagnose the failure and propose a refined cost function.

\begin{figure}[h!]
\centering
\fbox{\parbox{0.95\columnwidth}{
\small
\texttt{--- Outer Loop Triggered: Analyzing failure and refining strategy... ---}
\newline
\newline
\textbf{LLM Explanation:} The previous cost function was failing because it did not sufficiently prioritize the critical aspects of the task, such as ensuring the box is flipped onto its side and maintaining contact with the wall. The rewards and penalties were not balanced effectively to guide the robot towards achieving the task goal. Specifically, the orientation error was not heavily penalized, and the rewards for contact and elevation were not strong enough to drive the desired behavior.
\newline
\newline
In the new version, the following improvements have been made:
\begin{enumerate}
    \item \textbf{Increased Penalty for Orientation Error}: The weight on the angle error has been increased to emphasize the importance of achieving the correct orientation.
    \item \textbf{Stronger Rewards for Contact and Elevation}: The rewards for maintaining contact with the wall and elevating the box have been increased, encouraging the robot to push the box against the wall and attempt a flip.
    \item \textbf{Adjusted Proximity Penalties}: The penalties for the end-effector's distance from the box and the box's distance from the wall have been adjusted to better guide the robot's actions.
    \item \textbf{Relaxed Orientation Threshold}: The threshold for the orientation reward has been slightly relaxed to allow for some tolerance in achieving the target orientation, which can help in stabilizing the box on its side.
\end{enumerate}
\texttt{These changes aim to create a more balanced and effective cost function that better aligns with the task objectives. Simulated environment's cost function has been updated.}
}}
\caption[Example of LLM-driven failure diagnosis and refinement]{Direct output from the LLM during an online adaptation cycle. The model provides a clear, human-readable diagnosis of the previous plan's failure and details the specific changes made to the cost function for the next attempt.}
\end{figure}

\section{Example of an LLM-Generated Cost Function}
\label{app:cost_function_example}

Listing ~\ref{fig:cost_function_code} shows a direct example of a Python cost function generated by the `LLM (Task Formulation)' module for the multi-stage ``Push and Pick Cutting Board" task (T1). This code serves as the running cost, $q(x_t, u_t)$, within the MPPI objective (Eq.2). The function demonstrates the LLM's ability to translate a complex, sequential goal—``push the board until the handle is over the edge, then pick it"—into a structured, computable objective that the planner can optimize.

The function skillfully balances multiple, often competing, objectives by implementing a dynamic, state-aware logic:
\begin{itemize}
    \item \textbf{Dynamic Goal Blending:} It calculates a `grasp\_score' based on the board's stability and how far its handle overhangs the table edge. This score is used in a sigmoid function to create a ``soft switch" that smoothly transitions the robot's objective from pushing the board to picking the handle.
    \item \textbf{Phase-Specific Rewards:} The cost function provides rewards tailored to the current phase. In the push phase, it rewards progress in moving the board to the edge (`Push progress reward'). In the pick phase, it rewards correct gripper alignment and lifting the board (`Pick-phase alignment', `Reward lift after grasp').
    \item \textbf{Stability Constraints:} Throughout the push phase, it heavily penalizes any lateral drift or rotation of the board, ensuring a stable push.
    \item \textbf{Contact Management:} It explicitly manages contact, penalizing the gripper for losing contact with the board during the push phase.
\end{itemize}
This example highlights how CoRAL's LLM grounds abstract, sequential instructions into a sophisticated, mathematical objective that enables the execution of complex, long-horizon tasks.

\begin{lstlisting}[ style=mystyle, caption={The complete, LLM-generated running cost function $q(x,u)$ for the ``Push and Pick Cutting Board" task (T1). It demonstrates a sophisticated, multi-stage logic with a soft switch to blend objectives between the pushing and picking phases.}, label={fig:cost_function_code}]
def state_cost(self):
    # --- Helper lambdas ---
    def sigmoid(x): return 1.0 / (1.0 + np.exp(-x))
    def clip01(v):  return max(0.0, min(1.0, float(v)))

    # --- EE pose ---
    ee_state = p.getLinkState(self.panda, self.grasp_link,
                              computeForwardKinematics=True,
                              physicsClientId=self.cid)
    ee_pos = np.array(ee_state[0])
    cost = 0.0

    # --- Board pose/vel ---
    board_pos, board_ori = p.getBasePositionAndOrientation(self.board, physicsClientId=self.cid)
    board_pos = np.array(board_pos)
    linvel, angvel = p.getBaseVelocity(self.board, physicsClientId=self.cid)

    # --- World/task params with safe defaults ---
    push_axis = np.array(getattr(self, ``push_axis", [1.0, 0.0, 0.0]), dtype=float)
    push_axis /= (np.linalg.norm(push_axis) + 1e-9)
    table_edge_x = getattr(self, ``table_edge_x", self.board_init_pos[0] + 0.25)
    desired_overhang = getattr(self, ``desired_overhang", 0.06)

    # --- Contact strategy points ---
    pick_point  = np.array(getattr(self, ``best_contact_point", board_pos), dtype=float)
    push_point  = np.array(getattr(self, ``push_contact_point",
                                     board_pos + 0.5 * desired_overhang * push_axis), dtype=float)

    # --- Overhang & readiness for grasping ---
    overhang_m   = np.dot(pick_point - np.array([table_edge_x, board_pos[1], board_pos[2]]), push_axis)
    overhang_nrm = clip01(overhang_m / max(1e-6, desired_overhang))
    v = np.linalg.norm(linvel) + 0.5 * np.linalg.norm(angvel)
    stability = clip01(np.exp(-3.0 * v))
    grasp_score = 0.7 * overhang_nrm + 0.3 * stability
    r = sigmoid(12.0 * (grasp_score - 0.55))  # soft switch: r -> 1 as grasp becomes viable

    # --- Blended target following (Push vs. Pick) ---
    w_follow_push = 12.0 * (1.0 - r)
    w_follow_pick = 20.0 * r
    cost += w_follow_push * np.linalg.norm(ee_pos - push_point)
    cost += w_follow_pick * np.linalg.norm(ee_pos - pick_point)

    # --- Push progress reward ---
    cost -= 18.0 * (1.0 - r) * overhang_nrm

    # --- Contact management (for push phase) ---
    in_contact = bool(getattr(self, ``in_contact", True))
    cost += (22.0 * (1.0 - r)) * (0.0 if in_contact else 1.0)

    # --- Pick-phase alignment ---
    ee_quat = ee_state[1]
    if hasattr(self, ``handle_desired_ori") and self.handle_desired_ori is not None:
        q_diff = p.getDifferenceQuaternion(ee_quat, self.handle_desired_ori)
        w = max(-1.0, min(1.0, float(q_diff[3])))
        angle = 2.0 * math.acos(w)
        cost += (28.0 * r) * angle

    # --- Lateral drift & rotation penalties (for push phase) ---
    dp = board_pos - np.array(self.board_init_pos)
    axial  = np.dot(dp, push_axis) * push_axis
    lateral = dp - axial
    cost += 10.0 * (1.0 - r) * np.linalg.norm(lateral)
    q_diff_b = p.getDifferenceQuaternion(board_ori, self.board_init_ori)
    wb = max(-1.0, min(1.0, float(q_diff_b[3])))
    ang_b = 2.0 * math.acos(wb)
    cost += 8.0 * (1.0 - r) * ang_b

    # --- Reward lift after grasp ---
    lift_h = board_pos[2] - self.board_init_pos[2]
    cost -= (32.0 * r) * max(0.0, float(lift_h))

    # --- General penalties ---
    if hasattr(self, ``last_action"):
        cost += 0.001 * np.linalg.norm(self.last_action) ** 2
    cost += 0.01 * getattr(self, ``current_step", 0)

    return cost
\end{lstlisting}

\section{Example of an Incomplete Stage-Transition Condition}
\label{app:stage_transition_failures}

Multi-stage contact-rich tasks require the LLM-generated objective to decide not only what each subgoal is, but also when the planner should switch from one subgoal to the next. In our implementation, these switches are usually represented either as hard predicates or as soft progress scores that blend phase-specific costs. While this structure makes long-horizon tasks tractable, it also introduces a failure mode: the LLM may generate a stage-transition condition that is under-specified for the manipulation task.

We illustrate this issue using T1, the ``Push and Pick Cutting Board'' task. A successful execution requires a pre-grasp stage in which the board is pushed until the intended grasp region sufficiently overhangs the table. Only then should the planner transition to the picking stage. Thus, the transition condition should verify that the graspable region is exposed by a sufficient margin.

An example of an \emph{incomplete} transition condition is:
\begin{equation}
    d_{\mathrm{edge}} > \tau,
\end{equation}
where $d_{\mathrm{edge}}$ is the signed distance between the cutting-board edge and the table edge along the intended push direction selected by the LLM, and $\tau$ is a small threshold. Although this condition captures the idea of exposing the board beyond the table, it can be incomplete if $\tau$ is too small or if the measured edge does not correspond to the intended grasp region. In such cases, the planner may switch to the picking phase when only a small portion of the board is exposed, causing the end-effector to move toward a grasp pose before there is enough clearance for a stable grasp.

A more complete condition should require a task-specific minimum overhang and, ideally, board stability:
\begin{equation}
    d_{\mathrm{handle}} > \tau_{\mathrm{grasp}}
    \quad \text{and} \quad
    \|\dot{x}_{\mathrm{board}}\| < \epsilon_v,
\end{equation}
where $d_{\mathrm{handle}}$ measures the overhang of the intended grasp region, $\tau_{\mathrm{grasp}}$ is the minimum pickable overhang, and $\epsilon_v$ prevents transition while the board is still sliding.

\section{VLM Prompting for Physical Parameter Estimation}
\label{app:vlm_prompt}

This section details the query sent to the VLM (GPT-4o) to infer the physical properties of objects in the scene, which are used to parameterize the internal world model, $\theta$. Unlike methods that use VLMs for identification, our approach leverages the VLM's physical commonsense reasoning. The VLM's task is not to identify objects, but to estimate their unobservable physical attributes based on their known identity and estimated geometric state.

The prompt, shown in Listing~\ref{fig:vlm_prompt_example}, provides the model with all available context: the full scene image (passed implicitly to the multimodal model), the natural language task description $T$, and a JSON object for each relevant item. This JSON object is populated with the object's known semantic `label' (from its 3D model) and its current 6-DoF `pose', as estimated by FoundationPose.

The VLM's sole task is \textbf{Estimation}: It must use its vast, pre-trained knowledge about the real world to estimate the physical properties of the object, such as its `mass\_kg' and `friction\_coeff', based on its visual appearance (e.g., material, size) and the provided context. By constraining the output to be only the completed JSON object, we ensure the response is directly parsable by our system.

    \begin{lstlisting}[ style=mystyle, caption={The structured prompt sent to the multimodal model (GPT-4o) to act as our VLM. The model is provided with the object's known label and estimated pose, and is tasked only with filling in the unknown physical parameters (?).}, label={fig:vlm_prompt_example}]
# Prompt sent to GPT-4o to act as the VLM
You are a robotics expert with a deep understanding of physics.
Your task is to estimate the physical properties of an object for a simulation, based on its appearance in an image.

Task Description: ``Push the cutting board until the handle is off the table, then pick it up."

I have an object in the scene. I know what it is and I have an estimate of its current pose. Please provide your best estimate for its mass (in kg) and friction coefficient based on the image.

Object Data:
{
  ``cutting_board": {
    ``pose_estimated": [0.5, 0.0, 0.0, 0.0, 0.0, 0.0, 1.0],
    ``mass_kg": "?",
    ``friction_coeff": "?"
  }
}

Respond ONLY with the completed JSON object, filling in the unknown values.

# ----------------------------------------------------
# Example VLM JSON Response for the task above:
{
  ``cutting_board": {
    ``pose_estimated": [0.5, 0.0, 0.0, 0.0, 0.0, 0.0, 1.0],
    ``mass_kg": 0.4,
    ``friction_coeff": 0.4
  }
}
    \end{lstlisting}

\section{LLM Prompting for Cost Function Generation}
\label{app:cost_generation_prompts}

\begin{lstlisting}[style=mystyle,caption={General task-adaptive prompt used by the LLM to generate initial MPPI cost functions.},label={fig:llm_prompt_general}]
You are an expert in optimal control, physics-based planning, and contact-rich manipulation.
Your job is to generate an initial mppi cost function that adapts to the task description
and world parameters.

IMPORTANT: The structure must be GENERAL and TASK-ADAPTIVE.
Do NOT hard-code logic for a specific task. Instead, infer the requirements from the task description.

You must output ONLY a Python-like function:

    def state_cost(self):

STRUCTURAL REQUIREMENTS:
1. The cost must be composed of weighted terms for:
   - distance-to-goal or subgoal
   - contact or interaction constraints (if relevant)
   - orientation/alignment terms (if relevant)
   - force or stability terms (if relevant)
   - control effort
   - time/step penalty

2. If the task involves MULTIPLE PHASES (e.g., push then pick, flip then place),
   you MUST:
   - infer subgoals,
   - compute a phase progress score,
   - optionally blend objectives using a soft switch (sigmoid-based).

3. If the task involves CONTACT-RICH behavior (e.g., pushing, flipping, using a wall),
   include:
   - contact incentives or penalties,
   - drift or slippage penalties,
   - force-dependent shaping terms (if sensed).

4. If the task is SIMPLE (e.g., pick-and-place),
   use a single-stage goal-driven cost with alignment and distance terms.

5. ALL TERMS must be conditional on task semantics.
   Only include what is relevant for the input task.

6. ALL WEIGHTS must be numeric (choose reasonable magnitudes).

7. The function MUST be fully executable Python-like pseudocode using:
   - norms
   - dot products
   - quaternions (if needed)
   - sigmoid for soft transitions
   - optional heuristics (e.g., stability)

FORMAT (MANDATORY):
Return ONLY the code block:

    def state_cost(self):
        ...
        return cost

----------------------------------------
TASK DESCRIPTION:
"{TASK_DESCRIPTION}"

POSE_STATE:
{TRACKED_POSES_JSON}

PHYSICAL_PARAMS:
{ESTIMATED_PARAMS_JSON}

Return ONLY the Python code block.
\end{lstlisting}

\section{LLM Prompting for Online Adaptation}
\label{app:adaptation_prompts}

The `LLM (Online Adaptation)' module is triggered when the system detects persistent failures. Unlike the initial task formulation, the adaptation prompts are designed to be diagnostic, providing the LLM with a history of recent failed interactions to inform its corrections. The module employs two distinct prompting strategies depending on the type of refinement needed: strategy refinement (correcting the plan) and world model correction (correcting physical parameters).

\subsection{Strategy Refinement Prompt}
When the logic of the plan itself is suspected to be flawed, the system asks the LLM to act as a robotics programmer and rewrite the core `state\_cost' function. As shown in Listing~\ref{fig:cost_refinement_prompt}, the prompt provides the LLM with the task, the environment's attributes, the recent execution history (e.g., last 5 steps of object positions and resulting costs), and critically, the \textbf{current, failing source code} of the cost function. It is then instructed to return a corrected code block and a natural language explanation of its changes. This process is the source of the explainable failure recovery analysis presented in the main text.

    \begin{lstlisting}[ style=mystyle, caption={The Python function and prompt structure used for Strategy Refinement. The LLM is given the failing code and recent history to rewrite the cost function.}, label={fig:cost_refinement_prompt}]
# Python function that builds the prompt for cost function refinement.
def ask_state_cost_fn(self, task: str, history: list, env: BoxPushEnv):
    # Grab the current, failing source code of the cost function.
    try:
        current_src = inspect.getsource(env.state_cost.__func__)
    except (OSError, IOError):
        current_src = env.state_cost_src

    # --- The prompt sent to the LLM ---
    prompt = (
        f'`Task: {task}\n"
        ``You have full freedom to compute any cost that helps '{task}'.\n"
        ``History of last 5 steps (object_pos, resulting_cost):\n"
        f"<RECENT_EXECUTION_HISTORY>\n\n"
        ``Please output TWO things, separated by '---':\n"
        ``1) Python code for a method `def state_cost(self): ...` (indented block only)\n"
        ``2) A brief explanation why the previous cost was failing and how the new one addresses it.\n\n"
        f'`Here is the CURRENT implementation that is failing:\n"
        f"{current_src}"
    )
    
    # Query the LLM and parse the response (code_str, explanation)
    resp = openai.chat.completions.create(...)
    content = resp.choices[0].message.content
    code_str, explanation = content.split("---",1)
    return code_str.strip(), explanation.strip()
    \end{lstlisting}

\subsection{World Model Correction Prompt}
If the strategy is believed to be correct but the physical outcomes do not match the simulation (e.g., the robot pushes but the object barely moves), the system asks the LLM to act as a physicist and refine the object parameters. The prompt, detailed in Listing~\ref{fig:param_refinement_prompt}, provides the LLM with the agent's current belief about the physical parameters (mass, friction) and the recent execution history. The LLM's task is to analyze the discrepancy between actions and outcomes in the history and propose corrected physical parameters, returning them in a machine-parsable JSON format.

    \begin{lstlisting}[ style=mystyle, caption={The Python function and prompt structure for World Model Correction. The LLM analyzes the recent history to refine its belief about the object's physical properties.}, label={fig:param_refinement_prompt}]
# Python function that builds the prompt for physical parameter refinement.
def ask_params_refinement(self, history: list, object_params: list):
    
    # --- The prompt sent to the LLM ---
    prompt = (
        ``We have the following object parameters:\n"
        f"{json.dumps(object_params, indent=2)}\n\n"
        ``Recent execution history (last 5 steps):\n"
        f"<RECENT_EXECUTION_HISTORY>\n\n"
        ``Based on this, propose refined values for mass and friction_coef."
        ``Return ONLY a JSON array of objects with keys "
        "`label', `mass', and `friction'.\n"
        ``Example output:\n"
        "[\n"
        "  {\``label\":\``cutting_board\",\``mass\":0.45,\``friction\":0.35}\n"
        "]"
    )

    # Query the LLM and parse the JSON response
    resp = openai.chat.completions.create(...)
    text = resp.choices[0].message.content.strip()
    try:
        return json.loads(text)
    except json.JSONDecodeError:
        # Fallback to extract JSON if LLM adds extra text
        ...
    \end{lstlisting}

\section{LLM-driven Contact Strategy Generation}
\label{app:contact_strategy_generation}

A key challenge in contact-rich manipulation is determining precisely \textit{where} to make contact with an object. Uniformly sampling an object's entire surface is computationally inefficient and unlikely to yield strategically useful points. To overcome this, CoRAL leverages the LLM's commonsense physical reasoning to intelligently narrow the search space. This is achieved through a two-stage process, detailed in Listing ~\ref{fig:contact_code}, which translates a high-level task into a concrete set of candidate contact points ($C_0$).

\textbf{Stage 1: Strategic Region Proposal.} First, the `LLM (Task Formulation)' module queries a foundation model (GPT-4o) with a structured prompt that includes the task description and the VLM-estimated object parameters. The prompt, shown in Listing ~\ref{fig:contact_prompt}, explicitly instructs the LLM to act as a robotics expert and identify 1-3 small, promising surface regions for contact on each relevant object. The LLM is constrained to return this information in a structured JSON format, specifying each region's 3D center, surface normal, radius (extent), and the desired number of samples. For the ``Push and Pick Cutting Board" task, the LLM correctly identifies that pushing should occur on the board's main surface, while grasping should target the handle.

\textbf{Stage 2: Geometric Candidate Sampling.} Second, the structured JSON response from the LLM is passed to a geometric sampling function (`sample-points-in-region'). This function translates the LLM's abstract region definitions into a dense set of 3D point coordinates. For each region, it defines a 2D disk in 3D space oriented by the provided center and normal vectors. It then samples the requested number of points within this disk, generating the final set of candidate contact points, $C_0$, which are then used to bias the MPPI planner's exploration as described in the main text.

    \begin{lstlisting}[ style=mystyle, caption={The two-stage Python implementation for generating the contact strategy $C_0$. The `ask-region-strategy' function queries the LLM for high-level guidance, and `sample-points-in-region' translates that guidance into concrete 3D coordinates.}, label={fig:contact_code}]
# Stage 1: Query the LLM to propose strategic contact regions.
def ask_region_strategy(object_params, task_desc):
    # The prompt is shown in Figure~\ref{fig:contact_prompt}
    prompt = f"..." 
    resp = client.chat.completions.create(...)
    return resp.choices[0].message.content

# Stage 2: Sample concrete 3D points from an LLM-defined region.
def sample_points_in_region(region):
    c = np.array(region[``center"], dtype=float)
    n = np.array(region[``normal"], dtype=float)
    r = float(region[``extent"])
    k = int(region[``num_samples"])

    # Create two orthonormal tangent vectors to define the plane of the disk.
    if abs(n[2]) < 0.9:
        axis = np.array([0.0, 0.0, 1.0])
    else:
        axis = np.array([0.0, 1.0, 0.0])
    t1 = np.cross(n, axis)
    t1 /= np.linalg.norm(t1)
    t2 = np.cross(n, t1)
    t2 /= np.linalg.norm(t2)

    # Sample k points within the 2D disk defined by the tangents.
    pts = []
    for _ in range(k):
        rho   = np.sqrt(np.random.rand()) * r  # Uniform sampling in a disk
        theta = np.random.rand() * 2 * np.pi
        offset = rho * np.cos(theta) * t1 + rho * np.sin(theta) * t2
        pts.append((c + offset).tolist())
    return pts
    \end{lstlisting}

    \begin{lstlisting}[ style=mystyle, caption={The structured prompt and an example JSON response for the ``Push and Pick Cutting Board" task. The prompt constrains the LLM to provide a structured, machine-readable output.}, label={fig:contact_prompt}]
# The prompt sent to the LLM (Task Formulation) module:
Task: Push the cutting board until the handle is off the table, then pick it up.

You have objects with parameters:
{

  ``board": [{TRACKED_POSES_JSON} , {ESTIMATED_PARAMS_JSON}],
  ``table": [{TRACKED_POSES_JSON} , {ESTIMATED_PARAMS_JSON}]
}

Instead of uniform sampling, identify for each object 1-3 small surface regions
where contact is most promising.
For each region, return:
  - center: [x,y,z]
  - normal: unit surface normal [nx,ny,nz]
  - extent: radius in meters around center
  - num_samples: how many points to sample there

Respond ONLY a JSON mapping each label to its ``regions" array.

# Example LLM JSON Response:
{
  ``object": {
    ``regions": [
      {
        ``center": [0.5, 0.0, 0.01],
        ``normal": [0, 0, 1],
        ``extent": 0.1,
        ``num_samples": 30
      }
    ]
  }
    \end{lstlisting}

\end{document}